%% file: main.tex
\documentclass[10pt,twocolumn,letterpaper]{article}

\usepackage{iccv}              


\definecolor{iccvblue}{rgb}{0.21,0.49,0.74}
\usepackage[pagebackref,breaklinks,colorlinks,allcolors=iccvblue]{hyperref}

\definecolor{cvprblue}{rgb}{0.21,0.49,0.74}
\definecolor{nvidiagreen}{RGB}{119,185,0}
\usepackage{multirow}
\usepackage{comment}
\usepackage{xcolor}  
\usepackage{pifont}  
\usepackage{overpic}
\def\eg{\emph{e.g}\onedot} 
\def\ie{\emph{i.e}\onedot}

\def\etal{\emph{et al}\onedot}
\definecolor{orangeyellow}{RGB}{255,165,0}
\definecolor{darkyellow}{RGB}{255, 204, 0}

\newlength\savewidth\newcommand\shline{\noalign{\global\savewidth\arrayrulewidth
\global\arrayrulewidth 1pt}\hline\noalign{\global\arrayrulewidth\savewidth}}

\newcommand\haonote[1]{\textcolor{blue}{HAO: #1}}
\newcommand{\reffig}[1]{\cref{#1}}
\newcommand{\reftab}[1]{\cref{#1}}

\newcommand{\ui}[1]{\underline{\textit{#1}}}

\def\paperID{2492} 
\def\confName{ICCV}
\def\confYear{2025}

\title{Video2BEV: Transforming Drone Videos to BEVs \\ 
for Video-based Geo-localization}

\author{Hao Ju$^{1}$ \quad Shaofei Huang$^{1}$\quad  Si Liu$^2$ \quad Zhedong Zheng$^{1\dagger}$\\ 
{\small $^1$Faculty of Science and Technology and Institute of Collaborative Innovation, University of Macau}\\ {\small $^2$Institute of Artificial Intelligence, Beihang University}\\
{\tt\small \{yc47429,zhedongzheng\}@um.edu.mo}, {\tt\small shaofeihuang.ai@gmail.com}, {\tt\small liusi@buaa.edu.cn}}

\begin{document}
\maketitle
\renewcommand{\thefootnote}{\fnsymbol{footnote}} 
\footnotetext[2]{Corresponding author.}

\begin{abstract} 
Existing approaches to drone visual geo-localization predominantly adopt the image-based setting, where a single drone-view snapshot is matched with images from other platforms.
Such task formulation, however, underutilizes the inherent video output of the drone and is sensitive to occlusions and viewpoint disparity.
To address these limitations, we formulate a new video-based drone geo-localization task and propose the Video2BEV paradigm.
This paradigm transforms the video into a Bird's Eye View (BEV), simplifying the subsequent \textbf{inter-platform} matching process. 
In particular, we employ Gaussian Splatting to reconstruct a 3D scene and obtain the BEV projection. Different from the existing transform methods, \eg, polar transform, our BEVs preserve more fine-grained details without significant distortion.
To facilitate the discriminative \textbf{intra-platform} representation learning, our Video2BEV paradigm also incorporates a diffusion-based module for generating hard negative samples. To validate our approach, we introduce UniV, a new video-based geo-localization dataset that extends the image-based University-1652 dataset. 
UniV features flight paths at $30^\circ$ and $45^\circ$ elevation angles with increased frame rates of up to 10 frames per second (FPS).
Extensive experiments on the UniV dataset show that our Video2BEV paradigm achieves competitive recall rates and outperforms conventional video-based methods. Compared to other competitive methods, our proposed approach exhibits robustness at lower elevations with more occlusions.
The code is available at: 
\href{https://github.com/HaoDot/Video2BEV-Open}{https://github.com/HaoDot/Video2BEV-Open}.
\end{abstract}

\begin{figure}[!t]
    \centering
    \includegraphics[width=\linewidth]{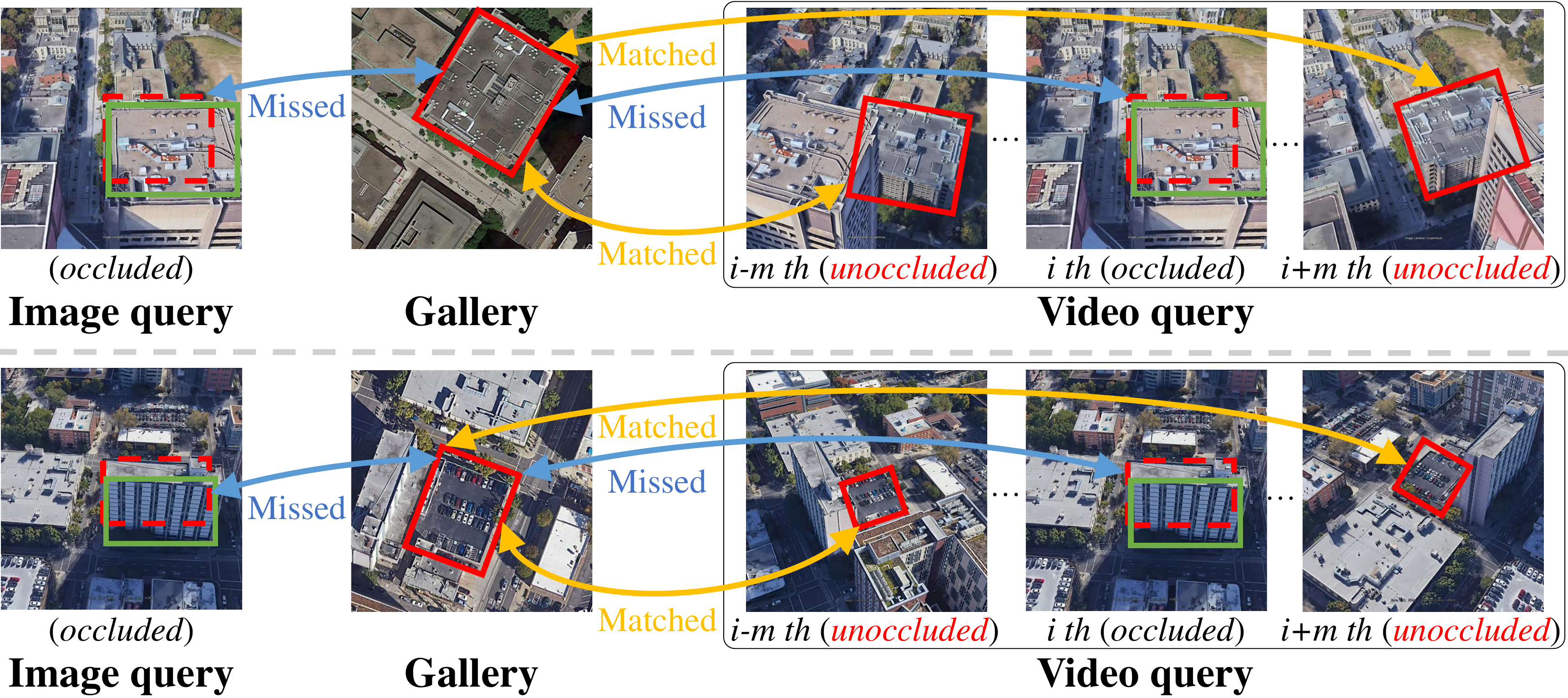}
    \vspace{-0.16in}
    \caption{
    \textbf{Typical failure cases for image-based drone geo-localization.}
    {
    For image queries (\emph{left}), the \textcolor{red}{core areas} in ground-truth are occluded by another building, largely compromising the spatial matching. In contrast, video queries  (\emph{right}) usually contain unoccluded frames in a circling flight, and thus could reflect a more comprehensive view of the target location. 
    }
    }
    \label{fig:vieo_better_img}
    \vspace{-0.25in}
\end{figure}
\section{Introduction}
Drone visual geo-localization aims to retrieve images of the same location from another platform, such as satellite, using visual information captured by the drone. This process is typically supported by off-line GNSS metadata~\cite{zheng_1652_mm20}, enabling drones to self-localize even in GNSS-denied environments, such as urban canyons or rural areas. 
The prevailing approach follows an \textbf{image-based} matching paradigm~\cite{zheng_1652_mm20,wang_dwdr_arxiv22,wang_lpn_tcsvt21,deuser_sample4geo_iccv23,dai_fsra_tcsvt21,mi_congeo_eccv24}, where a single drone-captured snapshot serves as the query to retrieve the corresponding location from the satellite-viewed candidate pool.
However, despite the advancements in image-based paradigms, two critical limitations persist. Both are due to drone flight height regulations~\cite{tran_management_ss22,bhat_autonomous_rco,konert_very_irs24}.
\textbf{(1)} Drones typically operate at lower altitudes in cluttered environments, resulting in significant occlusions in the captured images from buildings, trees, and other foreground objects.
Such occlusion can lead to a substantial loss or degradation of {visual information} in the drone image captured from a single viewpoint, making it difficult to establish accurate correspondences with satellite imagery.
As shown in~\reffig{fig:vieo_better_img} (\emph{left}), the core areas in the query are entirely obstructed by surrounding buildings. 
\textbf{(2)} Similarly, due to height limitation, drones typically capture images at oblique angles, while satellite images are predominantly acquired from a top-down perspective. 
The significant viewpoint disparity between the drone and satellite perspectives further increases the difficulty of matching.
\begin{figure}[t]
    \centering
    \includegraphics[width=\linewidth]{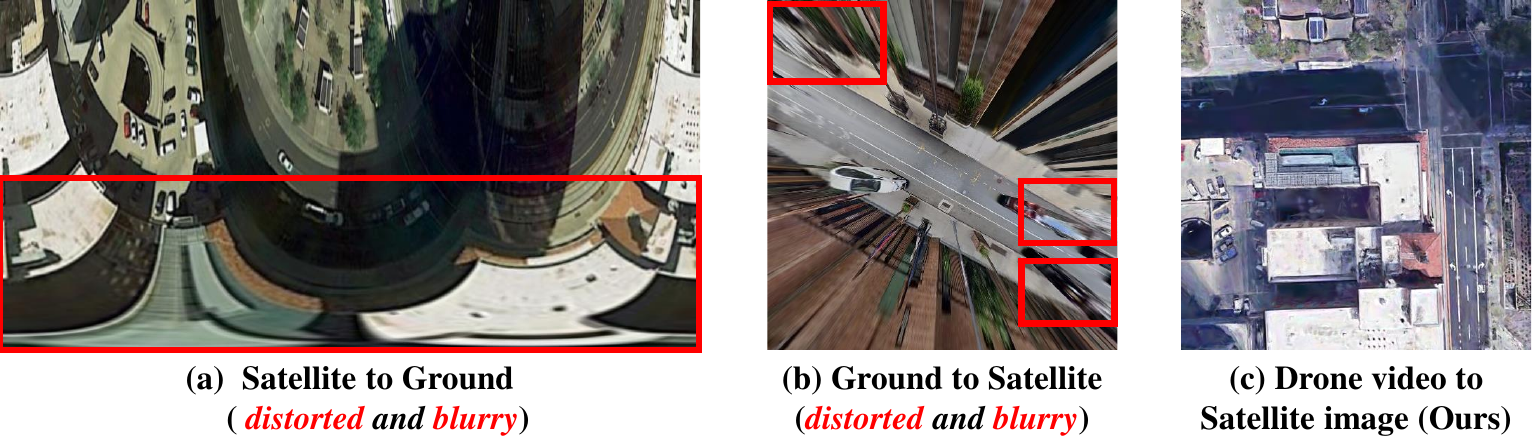}
    \vspace{-0.16in}
    \caption{
    Prevailing image-based geometric transformation (a) Satellite to Ground transformation by the polar transformation~\cite{shi_safa_nips19}, (b) Ground to Satellite transformation by the spherical transformation~\cite{wang_hcnet_nips24}. Our Drone video to Satellite transformation is shown in (c).
    Compared to image-based approaches, our method, fully leveraging the comprehensive view from the free-of-lunch drone videos, mitigates severe distortion and blurring.
    }
    \label{fig:comparison_alignment}
    \vspace{-0.22in}
\end{figure}

To address the limitations of the image-based geo-localization paradigm, we formulate a new \textbf{video-based} drone geo-localization task and propose a corresponding paradigm named \textbf{Video2BEV}, which leverages drone videos and transforms them into Bird’s-Eye View (BEV) representations for drone-satellite matching.
\textbf{(1)} Different from traditional single-view image approaches, our Video2BEV paradigm resorts to the multi-view nature of video to recover occluded regions and improve matching robustness.
As shown in~\reffig{fig:vieo_better_img} (\emph{right}), even though core areas of the target location are occluded in a certain drone-captured frame, we still can recover such core areas from other frames with different viewpoints. 
\textbf{(2)} In this work, we reconstruct the view in the BEV format, since BEV representation aligns with the satellite's top-down viewpoint, thus reducing the \textbf{inter-platform} discrepancy.
To convert input image into a calibrated format, existing methods usually apply 2D geometric transformations~\cite{shi_safa_nips19,wang_hcnet_nips24}, but suffer from spatial distortion and blurring (see~\reffig{fig:comparison_alignment}~(a, b)). 
Inspired by the success of 3D Gaussian Splatting (3DGS) in reconstruction~\cite{chabot_gaussianbev_arxiv24},  we introduce a new 3D-aware transformation. 
In particular, we leverage 3DGS to reconstruct the 3D scene based on the multi-view snapshots from the drone video, and then obtain the BEV representation via projection. 
As shown in~\reffig{fig:comparison_alignment}~(c), the BEV generated by our Video2BEV transformation exhibits fine-grained textures with minimal distortion or blurring, thus facilitating the subsequent \textbf{inter-platform} matching.
Furthermore, considering the nearby location with a similar visual appearance, the proposed Video2BEV paradigm further incorporates a diffusion-based hard negative synthesis module. This module generates BEV representations {that retain original semantic content but with different fine-grained discrepancies}, serving as hard negatives during training. 
By incorporating these challenging samples, the model learns to discriminate \textbf{intra-platform} samples from highly similar yet geographically distinct locations.

Finally, to support the video-based geo-localization task, we introduce a new dataset UniV with 2 drone videos per location accompanying with 16 ground-view snapshots and 1 satellite-view image, which is closer to the real-world deployment. With the help of unobstructed frames, the video input significantly reduces the impact of occasional occlusions present in the single image, thereby improving overall performance on all our re-implemented methods. For instance, LPN~\cite{wang_lpn_tcsvt21} receives +24.46\% AP increment (see ~\reffig{fig:video_better_img_performance}). 
In brief, our contributions are:

\begin{figure}[!t]
    \centering
    \includegraphics[width=\linewidth]{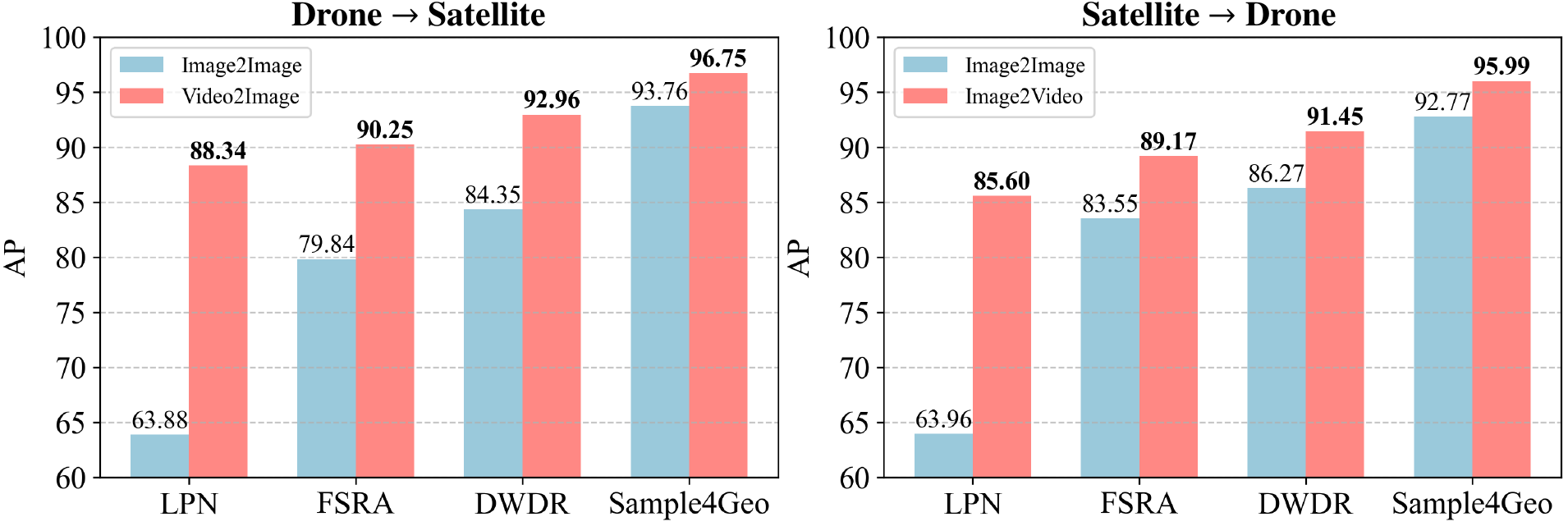}
    \vspace{-.16in}
    \caption{
    Performance comparisons of leveraging image data (Image2Image) or video data (Video2Image or Image2Video) with different methods including LPN~\cite{wang_lpn_tcsvt21}, FSRA~\cite{dai_fsra_tcsvt21}, DWDR~\cite{wang_dwdr_arxiv22}, and Sample4Geo~\cite{deuser_sample4geo_iccv23}. 
    We report the Average Precision (AP) metric.
    For a fair comparison, we keep the same number of data in the gallery. We could observe that all re-implemented methods achieve better performance when adopting video query or gallery.
    \label{fig:video_better_img_performance}
    }
    \vspace{-0.32in}
\end{figure}

\begin{itemize}
    \item We formulate a new video-based geo-localization task and propose the Video2BEV paradigm that transforms drone-view videos into BEV representations with the assistance of 3DGS, simplifying the subsequent \textbf{inter-platform} matching process. 
    To further enhance the \textbf{intra-platform} representation learning, we introduce a diffusion-based hard negative sample synthesis module, which generates challenging training samples to expand data diversity and improve discriminative capability.
    \item 
    To validate the video-based drone geo-localization task, we introduce a new dataset, called UniV, with 3,304 drone flight videos with corresponding satellite and ground-view images. Our experiments reveals two insights:
    (1) Video-based data shows consistent performance advantages over single-frame image retrieval across multiple metrics (see~\reffig{fig:video_better_img_performance}). 
    (2) The proposed Video2BEV achieves 96.80 AP on Drone Video $\rightarrow$ Satellite, outperforming other competitive methods. Furthermore, the trained model shows strong generalization capabilities, maintaining 91.50 AP when tested on the unseen real-world dataset, \ie, SUES-200, without fine-tuning.
\end{itemize}

\begin{table*}[!t]
    \centering
    \caption{
    (a) Dataset comparisons between UniV and other visual geo-localization datasets. G, S, and D denote ground-view, satellite-view, and drone-view, respectively.
    We enable video modality and add another common elevation angle of drone flight.
    (b) Elevation angles $\theta$ illustration. Top panel shows $\theta$ = $45^\circ$ and bottom panel displays $\theta$ = $30^\circ$. With a lower elevation angle, the new flight captures the target location with wider \textit{\textcolor[RGB]{0,176,240}{Field of View (FoV)}} but more \textit{\textcolor{red}{occlusions}}, thereby posing more challenges for drone visual geo-localization.
    }
    \vspace{-0.12in}
    \begin{minipage}{0.55\textwidth}
        \centering
        \subcaption{}
        \resizebox{\textwidth}{!}{
        \begin{tabular}{c|c|c|c|c}
        \shline
        Datasets & Platforms & \#data per location & Modality & Elevation \\
        \hline
        CVUSA~\cite{workman_cvusa_iccv15} & G, S  & 1 image + 1 image & Image & N/A \\
        Lin~\etal~\cite{lin_dataset_cvpr15} & G, S  & 1 image + 1 image & Image & $45^\circ$ \\
        Vo~\etal~\cite{vo_dataset_eccv16} & G, S  & 1 image + 1 image & Image & N/A \\
        Tian~\etal~\cite{tian_dataset_cvpr17} & G, S  & 1 image + 1 image & Image & $45^\circ$ \\
        CVACT~\cite{liu_cvact_cvpr19} & G, S  & 1 image + 1 image & Image & N/A \\
        Vigor~\cite{zhu_vigor_cvpr21} & G, S  & 2 images + 1 image & Image & N/A \\
        SUES-200~\cite{zhu_sues_tcsvt23} & S, D & (1 + 50 $\times$ 4) images & Image & $45^\circ\sim70^\circ$ \\
        University-1652~\cite{zheng_1652_mm20} & G, S, D & (16 + 1 + 54) images & Image & $45^\circ$ \\
        GeoText-1652~\cite{chu_geotext_eccv24} & G, S, D & (16 + 1 + 54) images + 180 texts & Image + Text & $45^\circ$ \\
        \hline
        UniV  & G, S, D & (16 +1) images + \pmb{2 videos} & Image +\textbf{Video} & \pmb{$30^\circ$}, $45^\circ$ \\
        \shline
        \end{tabular}%
        }
        \label{tab:dataset}
    \end{minipage}
    \begin{minipage}{0.4\textwidth}
        \centering
        \subcaption{}\vspace{-.05in}
        \includegraphics[width=0.95\textwidth]{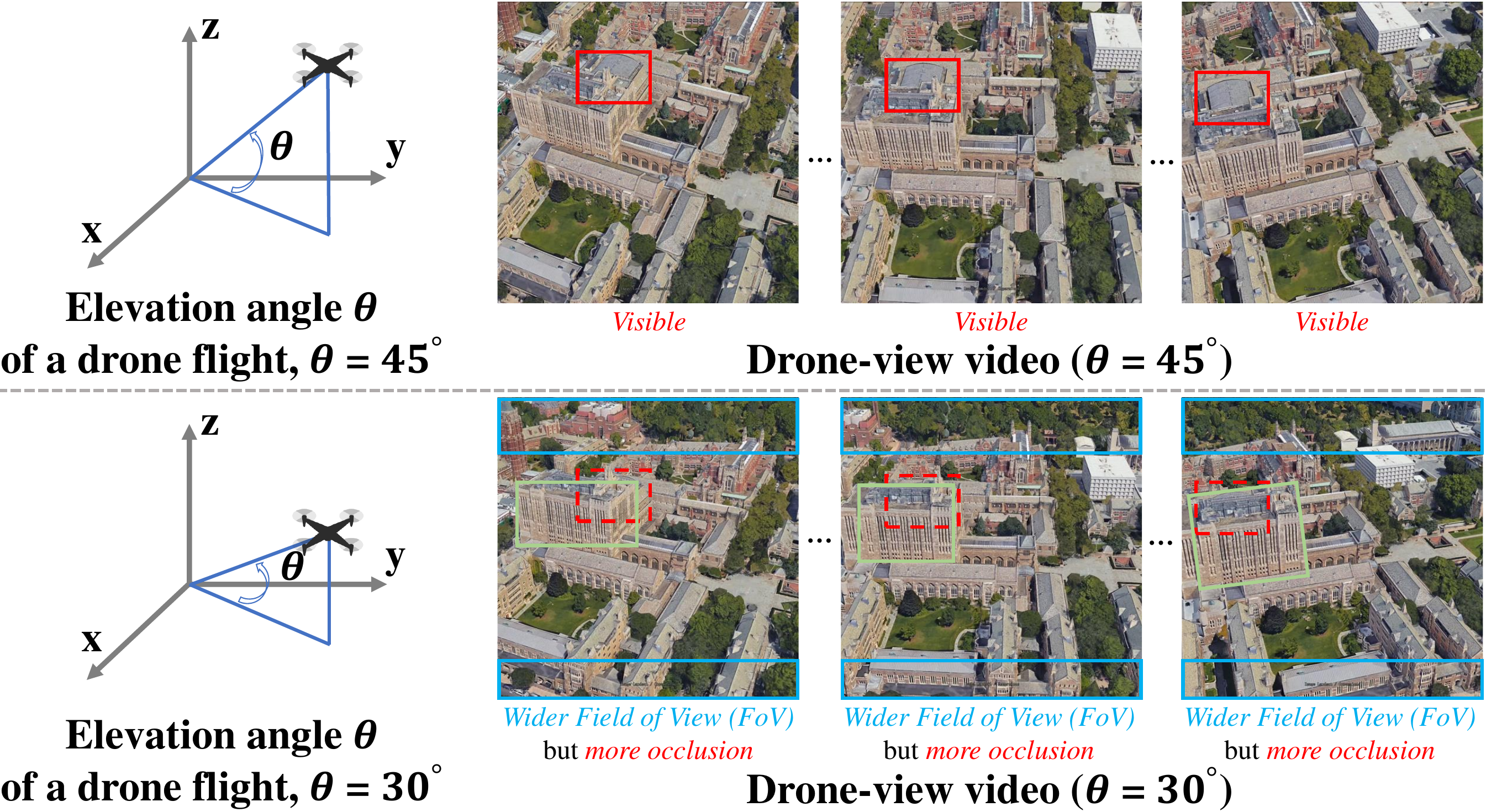}
        \label{fig:dataset}
    \end{minipage}
    \vspace{-0.15in}
\end{table*}

\section{Related Work}

\noindent\textbf{Image-based Geo-localization.} Image-based geo-localization, which is usually regarded as a sub-task of image retrieval, applies image query to determine locations~\cite{wilson_geo_survey_ijcv24,zhang_geodtr_pami24}. 
The primary challenge of this task is the large appearance discrepancy due to different viewpoints across platforms, including ground~\cite{huynh_contrastive_eccv24}, satellite~\cite{dutta_multiview_cvpr24}, and drone~\cite{ding_object_pami21,zhu_visdrone_pami21}. 
Previous methods can be coarsely divided into two families: image-level alignment and feature-level alignment. 
\textbf{(1) For image-level alignment}, 
Shi~\etal~\cite{shi_safa_nips19} leverage polar transformation to warp satellite images to the ground view. Similarly, Wang~\etal~\cite{wang_hcnet_nips24,ye_pano_bev_eccv24} transform ground images to the satellite view.
Regmi~\etal~\cite{regmi_gan_geo_iccv19} synthesize aerial images from ground images to facilitate matching with satellite view via Generative Adversarial Networks (GANs). 
Tian~\etal~\cite{tian_bev_gan_tcsvt21} employ GANs to transform drone-view images into satellite-view images. Andrea~\etal~\cite{vallone_danish_iral22} convert drone-view images to ground views through 3D reconstruction but the output is with distortions.
\textbf{(2) For feature-level alignment,} 
Dai~\etal~\cite{dai_fsra_tcsvt21} and Wang~\etal~\cite{wang_lpn_tcsvt21} establish feature alignment in a region-correspondence manner.
Some methods~\cite{lin_joint_tip22,song_unified_tip24} focus on key-point alignment.
Other methods aim to improve the discriminative ability of neural networks with tailored modules, such as 
lite-transformer encoder~\cite{wang_mfrgn_mm24}, layer-to-layer attention block~\cite{zhu_transgeo_cvpr22,yang_cross_nips21}, adaptive integration module~\cite{sun_tirsa_tcsvt24}, strong backbones~\cite{yang_dinov2_ral25}, adaptation information consistency module~\cite{li_unsupervised_tgrs24}, spatially-adaptive denormalization~\cite{wang_environment_pr24}
and 
well-designed loss functions, \eg, 
semantic augmentation loss~\cite{zhang_geodtr_aaai23}, contrastive loss~\cite{zhang_geodtr_pami24,mi_congeo_eccv24,li_unsupervised_geo_cvpr24,deuser_sample4geo_iccv23}, dynamic weighted decorrelation regularization~\cite{wang_dwdr_arxiv22}, peer learning~\cite{zeng_graphgeo_tmm22}, instance loss~\cite{zheng_1652_mm20} and the optimal transport~\cite{shi_cvft_aaai20}.
Additionally, other methods~\cite{regmi_video_iccv21,liu_cvact_cvpr19,shi_orientation_cvpr20} fuse the extra orientation meta-information from GNSS with extracted features.
However, previous methods overlook the viewpoint variation within drone-view videos, thinking in an image-based setting. Our method is among the early attempts to leverage the viewpoint variation of drone-view videos to transform drone-view video to Bird's-Eye View (BEV), thereby reducing the viewpoint disparity between drone and satellite views.

\noindent\textbf{Video-based Geo-localization}
While video understanding has been a focus of the computer vision community for decades, the problem remains challenging due to the complexity added by the time dimension and the volume of data.
Early works~\cite{carreira_i3d_cvpr17,shen_stepnet_tmm24,feichtenhofer_slowfast_cvpr19} leverage two-stream convolution networks to fuse spatial semantic information with motion information. 
Subsequently, attention mechanisms~\cite{darcet_register_arxiv23,han_vit_survey_pami22,peirone_backpack_cvpr24} have been introduced for long-term video understanding, such as vanilla self-attention~\cite{arnab_vivit_iccv21,vaswani_attention_nips17}, shift window~\cite{liu_swin_iccv21,liu_videoswin_cvpr22}, masked auto-encoder~\cite{he_mae_cvpr22,tong_videomae_nips22}, and local spatiotemporal attention~\cite{son_csta_cvpr24}.
Recently, large language models~\cite{tang_video_llm_arxiv23} have also shown their superiority in video understanding.
In visual geo-localization, videos contain more visual information captured through the camera's trajectory, which can provide more comprehensive information compared to images. 
Vyas~\etal~\cite{vyas_gama_eccv22} are the first to collect ground-view data in video format and propose a hierarchical approach for processing clips of ground-view videos.
Regmi~\etal~\cite{regmi_video_iccv21} leverage the geo-temporal proximity between the ground-view videos and GNSS locations to extract coherent features from videos. 
Expanding to a global scale, Kulkarni~\etal~\cite{kulkarni_cityguessr_eccv24} introduce a large-scale ground-view video dataset for worldwide geo-localization.
Different from the ground-view videos, drone-view videos typically contain multi-view and multi-scale information for the target location~\cite{mildenhall_nerf_eccv20}.
{In this paper, we collect drone-view data in video format and propose a video-based geo-localization dataset. Rather than adopting the off-the-shelf video backbone, we propose a Video2BEV transformation to leverage the 3D geometric correspondences and enable a straightforward spatial alignment for matching.} 

\section{The UniV Dataset}
Given the lack of a video-based drone geo-localization benchmark, we collect a new dataset dubbed \emph{UniV} involving the video modality. 
We follow the location information and the protocol of the existing University-1652 dataset~\cite{zheng_1652_mm20}.  
The UniV dataset encompasses 1,652 {locations} in 72 universities from three platforms, \ie, ground, satellite, and drone cameras. In particular, {the UniV dataset contains 16 ground-view images and 1 satellite-view image for each location} and the training set of UniV dataset contains 701 {locations}, while the test set in the UniV dataset includes other 951 {locations}. There are no overlapping locations between the training and test sets. 
The proposed UniV dataset is different from the image-based University-1652 and other datasets in two primary aspects, \ie, modality and elevation-angle expansions (see ~\reftab{tab:dataset}).

\noindent\textbf{Modality Expansion.} 
Existing datasets~\cite{workman_cvusa_iccv15,vo_dataset_eccv16,liu_cvact_cvpr19,zhu_vigor_cvpr21} collect data from two platforms, \eg, satellite and ground. 
Although some datasets~\cite{zheng_1652_mm20,tian_dataset_cvpr17,zhu_sues_tcsvt23} include drone views, collected data is still in an image format. 
We adopt similar operations as the University-1652 but collect drone-view data in a video format.
Specifically, we leverage the 3D {engine of} google earth~\cite{mutanga_google_earth_rs19} to simulate the real-world movement of a drone equipped with a camera.
To collect video data containing various scales and viewpoints, 
we leverage the dynamic viewpoints within the 3D {engine} and set the moving viewpoints along a spiral curve for moving around the target location in three circles, closely approximating real-world drone flights. 
All videos are collected in 30 frame rate. 
Considering the video redundancy, in practice, we subsample videos along the temporal dimension, resulting in frame rates of 2, 5, and 10 for further processing. 

\noindent\textbf{Elevation-angle Expansion.}
Conventional datasets~\cite{zheng_1652_mm20, lin_dataset_cvpr15, tian_dataset_cvpr17} collect drone data in a fixed elevation angle, \ie, $45^\circ$, {which does not fully simulate the real-world cases.} 
Therefore, we add one new synthetic flying path at another common setting, \ie, a lower elevation angle~$30^\circ$. 
The new flying path poses two new challenges (see~\reftab{fig:dataset}). 
First, drones flying at a $30^\circ$ elevation angle capture scenes that include the target location and more surrounding areas, providing a wider Field of View (FoV), thus introducing disruptions for the center target location during matching. 
Second, flight paths at a lower elevation angle lead to more occluded cases, which lay over the core areas of the target location. 
It poses challenges {for mining} the discriminative frames in the video, {whereas it becomes easier when captured at a $45^\circ$ elevation}.
Therefore, the proposed dataset could further evaluate the robustness of methods against more disruptions and occlusions, which is closer to real-world drone geo-localization usage.

\noindent\textbf{Discussion. The contribution to the community.}
Different from existing datasets~\cite{zheng_1652_mm20, lin_dataset_cvpr15, tian_dataset_cvpr17}, the proposed UniV expands the modality from image to video (see~\reftab{tab:dataset}), facilitating the development of robust drone visual geo-localization.
A single image provides limited information about the corresponding location. When core areas of the location are occluded, single-image queries can not produce reliable matching results (see~\reffig{fig:vieo_better_img}). In such cases, the video contains both occluded and unoccluded frames. One frame may contain core-area information to complement another frame and together they can provide robust and complete information required for drone visual geo-location. In this way, all re-implemented methods perform better when adopting video data (see~\reffig{fig:video_better_img_performance}).
Moreover, the UniV dataset 
{also introduces a new real-world challenge}
for drone visual geo-localization. The new elevation angle of $30^\circ$ is 
{typical} in real-world 
{flights}\footnote{The United States and the United Kingdom allow drone flights up to 400 feet; China restricts drones up to 120 meters.}. The $30^\circ$ elevation angle {faces} more occlusion cases (see~\reftab{fig:dataset}), simulating outputs of real-world drone flights.

\section{Method}
\subsection{Video2BEV Transformation}
\label{sec:bev_transformation}
\begin{figure*}[!t]
    \centering
    \includegraphics[width=\linewidth]{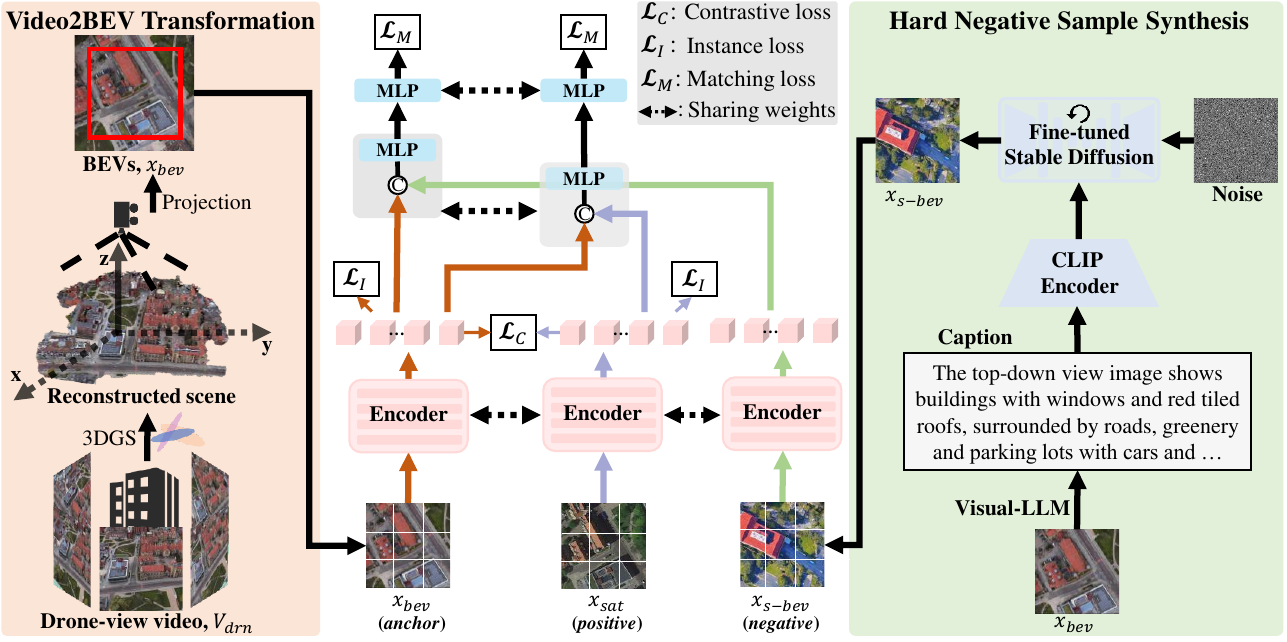}
    \vspace{-0.26in}
    \caption{
    {The overview of the Video2BEV paradigm. \textbf{Video2BEV Transformation (\textit{left}).}} Given drone-view video {$V_{drn}$} containing multi-view frames, we adopt 3D Gaussian Splatting~(3DGS) to reconstruct the scene at first. Then we render the scene from a Bird-Eye-View to get the {projection} (BEVs). {Considering the region of the core area, we further crop BEVs for training. We can observe that} BEVs exhibit resemblances to the corresponding satellite-view images. 
     \textbf{Hard Negative Sample Synthesis (\textit{right}).} 
     Given captions generated by an {off-the-shelf} visual-LLM~\cite{hu_minicpm_arxiv24}, we fine-tune a stable-diffusion model~\cite{rombach_stable_diffusion_cvpr22} with LoRA~\cite{hu_lora_iclr22}, and conduct inference to synthesize samples which serve as negative samples for subsequent usage.
    \textbf{Model Architecture (\textit{middle}).} 
    Given outputs of the proposed Video2BEV transformation, we extract embeddings by a shared encoder for satellite images $x_{sat}$ and BEVs $x_{bev}$, supervised by the contrastive loss $\mathcal{L}_{C}$ and the instance loss $\mathcal{L}_{I}$. 
    Then we extract embeddings from synthetic BEVs $x_{s\text{-}bev}$ and adopt {MLP} to fuse both positive and negative samples, supervised by the matching loss $\mathcal{L}_M$.
    Similar operations for satellite-view images are omitted.
    }
    \label{fig:architecture}
    \vspace{-0.05in}
\end{figure*}
\begin{figure}[!t]
    \centering
    \includegraphics[width=\linewidth]{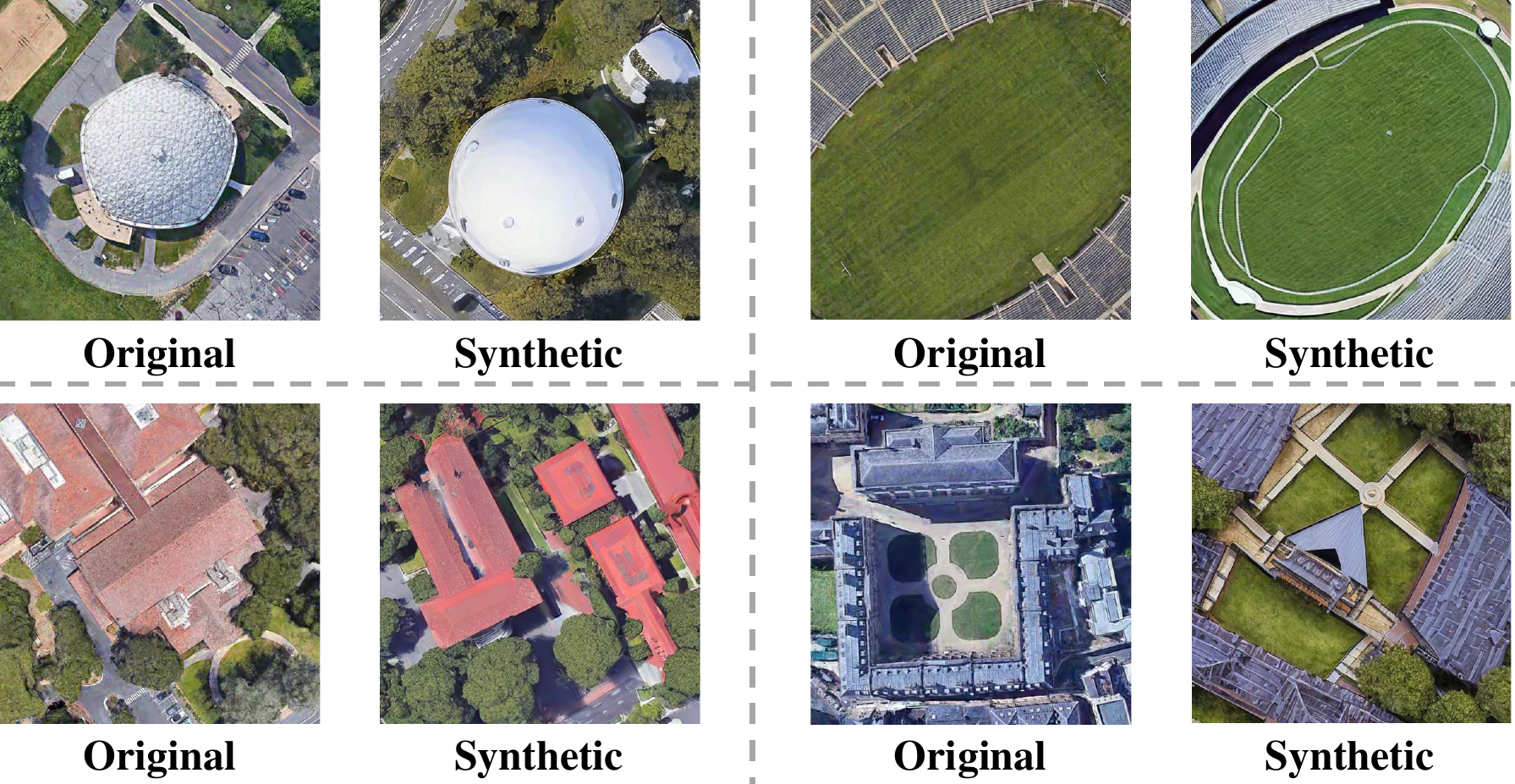}
    \vspace{-0.26in}
    \caption{
    Visualizations of original images and synthetic {hard negatives}.
    Synthetic negatives exhibit similar colors and structures to original images, which assures the quality of negatives.
    }
    \label{fig:ddgs_visualization}
    \vspace{-0.15in}
\end{figure}
During the flight around the target location, the viewpoints of the camera change dynamically, resulting in captured drone-view videos that contain rich multi-view information about both the target location and surrounding areas.
We explicitly leverage the multi-view information and transform the drone-view video into Bird's Eye Views (BEVs). In doing so, we ease the learning process for the subsequent model. Instead of learning geometry correspondence and feature correspondences simultaneously~\cite{chabot_gaussianbev_arxiv24}, the subsequent model only needs to learn the feature mapping relationship between two views, thus significantly facilitating network convergence.
As shown in the left part of~\reffig{fig:architecture}, given the drone-view video containing multi-view images, we estimate corresponding camera poses by structure from motion~\cite{ullman_sfm_1979} and reconstruct the scene containing the target locations utilizing 3D Gaussian Splatting (3DGS)~\cite{kerbl_3dgs_tg23}.
After reconstructing the scene, we adopt the normalized camera pose and the unit vector in the world coordinate to calculate the BEV camera pose and render BEVs. 
In particular, the vanilla 3DGS takes less than 8 seconds to render 50 BEVs with the shape of 512 $\times$ 512 on a NVIDIA 4090 GPU. 
Since we usually foreknow the search area in practice, the BEV generation process can be regarded as an off-line pre-processing. 
BEVs rendered with our Video2BEV transformation module do not suffer from severe distortion (see ~\reffig{fig:architecture}), thereby aiding in the subsequent \textbf{inter-platform} matching process with the satellite images.

\subsection{Hard Negative Sample Synthesis}
\label{sec:DDGS}
Negative samples play a significant role in discriminative metric learning.
Current negative sample mining strategies~\cite{deuser_sample4geo_iccv23,mi_congeo_eccv24,zhang_geodtr_pami24} cannot ensure the quantity and quality of negative samples, {as challenging samples are inherently scarce, and the selected negative samples do not necessarily exhibit similar architectural styles or consistent semantic details as the original samples.}
In order to bypass these drawbacks, we propose {to generate diverse BEV representations as hard negative samples through a fine-tuned diffusion model},
which is shown in the right of~\reffig{fig:architecture}.
After transforming the drone-view video to BEVs $x_{bev}$ via our Video2BEV transformation, we utilize a visual-LLM~\cite{hu_minicpm_arxiv24} to generate captions for both BEV and satellite-view images.
After obtaining captions for the BEVs and satellite images, we fine-tune a stable diffusion network~\cite{rombach_stable_diffusion_cvpr22} with LoRA~\cite{hu_lora_iclr22} to generate diverse synthetic images, during which we freeze the CLIP text encoder~\cite{radford_clip_icml21}.
The outputs of this model are negative samples for the subsequent step.
Since the transformed BEV and satellite images share the same top-down viewpoint and semantic content, we conduct inference on the same diffusion model to generate negative samples for both BEV and satellite-view images {using corresponding captions}.
We provide visualizations of original and synthetic images in~\reffig{fig:ddgs_visualization}. 
{Synthetic negative samples exhibit similar semantic contents as the original ones but with different fine-grained information, thus enhancing the discriminative \textbf{intra-platform} representation learning}.

\subsection{Model Optimization}
We adopt a general architecture from vision-language models~\cite{li_albef_nips21,li_blip_icml22}, enhanced by BEVs and synthetic negative samples.
The model architecture is shown in the middle of~\reffig{fig:architecture} and is optimized in a \textbf{two-stage} manner following~\cite{li_blip_icml22,zeng_xvlm_icml22}.
In the \textbf{first stage}, we transform the drone-view video to BEVs by the proposed Video2BEV transformation (see the left part of~\reffig{fig:architecture}). Then, we adopt a shared encoder to extract embeddings from paired BEV and satellite-view images. The encoder is ViT-S~\cite{dosovitskiy_vit_iclr21} excluding the classifier.
The supervisions of this stage are the instance loss $\mathcal{L}_I$~\cite{zheng2017dual} with the square-ring partition~\cite{wang_lpn_tcsvt21} and the contrastive loss $\mathcal{L}_C$~\cite{li_albef_nips21}. 
We apply multiple classifier modules to each part of the embeddings (similar to LPN~\cite{wang_lpn_tcsvt21}), yielding the location probability of two views which are denoted as $\hat{p}_{sat}$ and $\hat{p}_{bev}$ respectively.
The instance loss $\mathcal{L}_I$ {is formulated as the location classification} as :
\begin{equation}
    \mathcal{L}_I= -log({\hat{p}_{sat}}) -log({\hat{p}_{bev}}).
\end{equation}
Then we accumulate instance losses from multiple parts to form the final instance loss.
For the contrastive loss, given a pair of satellite-view and BEV images, the satellite-to-BEV similarity is defined as:
\begin{equation}
\label{eq:similarity_score}
    {S}_{sat2bev}=\frac{exp(s(f_{sat},f_{bev})/\tau)}{\Sigma^N_{j=1}exp(s(f_{sat}, f^j_{bev})/\tau)},
\end{equation}
where $f_{sat}$ and $f_{bev}$ are embeddings of the same location from two platforms, and $f^j_{bev}$ denotes the sample within the mini-batch.
$\tau$ is a learnable temperature parameter.
$s(\cdot,\cdot)$ denotes the cosine similarity.
Similarly, the BEV-to-satellite similarity is $S_{bev2sat}$ and the contrastive loss $\mathcal{L}_{C}$ is:
\begin{equation}
    \mathcal{L}_{C} = -\frac{1}{2}(log(S_{sat2bev}) + log(S_{bev2sat})).
\end{equation}
In the \textbf{second stage}, we employ a two-layer MLP alongside the square-ring partition~\cite{wang_lpn_tcsvt21} to 
fuse two embeddings obtained from anchor-positive or anchor-negative samples
Specifically, when BEV serves as the anchor, satellite and synthetic BEV serve as the positive sample and the negative sample, respectively. Similarly, when satellite acts as positive samples, BEV and synthetic satellite act as positive and negative samples respectively.
Then we project the fused embeddings into the two-dimensional space using another MLP.
Given inputs from paired samples, the matching loss $\mathcal{L}_M$ between them is calculated as:
\begin{equation}
\begin{split}
    \mathcal{L}_M = -(
        p_{m}log(\hat{p}_{m})
        +(1-p_{m})log(1-\hat{p}_{m})
    ),
\end{split}
\end{equation}
where \( \hat{p}_m\) is the estimated matching probability and $p_{m}$ is a ground-truth binary label. If the two input data do not contain the synthetic data and are both from the same location, then \( p_{m} = 1 \); otherwise, \( p_{m} = 0 \).
Specifically, for the BEVs, we calculate the matching loss two times. 
For the first calculation, we rank the similarity $S$ and select {three} negative samples from the satellite-view images, ensuring that these samples do not belong to the same location simultaneously.
For the second calculation, we similarly select another three negative samples from the synthetic BEVs, {which are actually hard negatives from the same location.}
We apply similar operations to the satellite-view input.
Finally, we accumulate and average matching losses across different combinations of inputs.
In summary, the loss functions in our method include the instance loss $\mathcal{L}_I$, the contrastive loss $\mathcal{L}_C$, and the matching loss $\mathcal{L}_M$.
Specifically, {in the first stage of training, we optimize the encoder, classifier modules of our model, the temperature parameter $\tau$ with the instance loss $\mathcal{L}_I$ and the contrastive loss $\mathcal{L}_C$. Subsequently, we freeze the parameters fine-tuned in the first stage and train the MLPs from scratch in the second stage under the supervision of matching loss $\mathcal{L}_M$.}

\noindent\textbf{Discussion. What are the advantages of the synthetic negative samples?}
Inspired by successes in other fine-grained tasks~\cite{zheng2017unlabeled,shrivastava2017learning,zheng2019joint}, we encourage the model ``see'' more samples to prevent over-fitting as well as facilitate discriminative intra-platform feature learning. 
We are similar to GeNIe~\cite{koohpayegani_genie_arxiv23} in that both methods alter the image representation of the target object to generate hard negative samples. However, it is worth noting that there are two primary differences.
(1) We have a larger modification space, and the negative sample pool is no longer constrained to a fixed size. Different from the GeNIe in changing limited categories for classification, we perturb the initial noise of the diffusion model, and it leads to diverse generations. Utilizing the diffusion model, we can theoretically generate an infinite number of images as negative samples, expanding the negative sample pool significantly. 
(2) We retain the semantic content of the anchor samples in our synthetic negative samples. 
This is because we employ identical captions from the original samples to synthesize the negative samples while only changing the initial noise. This slight modification ensures that our negative samples are appropriately challenging, and encourages the model to check the fine-grained discrepancies among semantical-similar samples.

\section{Experiment}
\noindent\textbf{Implementation Details.}
Since the captions for the two views (drone and satellite) employ different wording to describe the same location, we synthesize samples based on the text and generate separate negative samples for each view.
All input images are resized to 256$\times$256.
We train the first stage of the proposed model with the AdamW optimizer, with a batch size of 140, for 140 epochs, and a learning rate of $2e^{-5}$ and $2e^{-4}$ for the encoder and other modules in the first stage respectively.
Then we freeze parameters in the first stage and train the second stage from scratch with a similar training configuration.
During the test stage, we utilize the similarity scores from the first stage to select the top 32 samples from the gallery, and then re-rank these top 32 samples in the second stage.
More details are provided in the supplementary materials.

\noindent\textbf{Evaluation Metrics.}
Satellite-view data is in image format, while drone-view data is collected in video format.
We can treat drone view data as images or video.
In this paper, we adopt the video setting for the evaluation of competitive methods and our method. Specifically, we treat a drone video as an individual query or gallery by averaging the similarity scores of the images within the video in a late fusion manner. 
There is a similar averaging operation on similarity scores of the BEVs which is also in video format. 
\begin{table}[t]
  \centering
  \caption{(a) Comparisons on the UniV for geo-localization between Drone (D) and Satellite (S) platforms. R@1 is recall at top1. AP (\%) is average precision (high is good).
  (b) Comparisons in terms of an out-of-distribution testing on the SUES-200 ($45^\circ$ test set). Our method still yields the best results.
  }
  \vspace{-0.2in}
  \begin{tabular}{c}
    \begin{subtable}{0.46\textwidth}
      \centering
      \caption{}
      \resizebox{\textwidth}{!}{%
        \begin{tabular}{c|cc|cc|cc|cc}
          \shline
          \multirow{3}[4]{*}{\raisebox{2ex}{Method}} & \multicolumn{4}{c|}{$\theta=45^\circ$}       & \multicolumn{4}{c}{$\theta=30^\circ$} \\
          \cline{2-9} & \multicolumn{2}{c|}{D$\rightarrow$S} & \multicolumn{2}{c|}{S$\rightarrow$D} & \multicolumn{2}{c|}{D$\rightarrow$S} & \multicolumn{2}{c}{S$\rightarrow$D} \\
          & R@1 & AP    & R@1 & AP    & R@1 & AP    & R@1 & AP \\
          \hline
          LPN~\cite{wang_lpn_tcsvt21}  & {86.31} & 88.34 & 83.31  & 85.60 & 68.62 & 72.50 & 67.76 & 71.30 \\
          FSRA~\cite{dai_fsra_tcsvt21}  & 88.59 & 90.25 & 87.30  & 89.17 & 81.60 & 84.17 & 77.89 & 81.00 \\
          DWDR~\cite{wang_dwdr_arxiv22}  & {91.73} & 92.96 & 89.87  & 91.45 & 88.02 & 89.81 & 85.59 & 87.85 \\
          Sample4Geo~\cite{deuser_sample4geo_iccv23} & 96.29 & 96.75 & 95.29 & 95.99 & {83.02} & {86.00} & {80.45} & {82.68} \\
          Ours  & \textbf{96.29} & \textbf{96.80} & \textbf{96.01} & \textbf{96.57} & \textbf{91.73}   & \textbf{93.01} & \textbf{92.58}   & \textbf{93.65} \\
          \shline
        \end{tabular}}
        \label{tab:performance}%
    \end{subtable} \\[1em]
    
    \begin{subtable}{0.36\textwidth}
      \centering
      \vspace{-0.05in}
      \caption{}
      \resizebox{0.9\textwidth}{!}{%
        \begin{tabular}{c|cc|cc}
          \shline
          \multirow{2}[4]{*}{\raisebox{1.5ex}{Method}} & \multicolumn{2}{c|}{D$\rightarrow$S} & \multicolumn{2}{c}{S$\rightarrow$D} \\
          \cline{2-5} & R@1   & AP    & R@1   & AP \\
          \hline
          LPN & 41.25  & 49.05  & 18.75  & 26.35  \\
          FSRA & 48.75  & 54.64  & 32.50  & 40.09  \\
          DWDR & 71.25  & 75.02  & 70.00  & 74.60  \\
          Sample4Geo & 81.25  & 84.14  & 86.25  & 88.80  \\
          Ours & \textbf{89.74}  & \textbf{91.50}  & \textbf{91.25}    & \textbf{92.53} \\
          \shline
        \end{tabular}}
        \label{tab:realworld}%
    \end{subtable}
  \end{tabular}
  \vspace{-0.2in}
  
\end{table}

\begin{table*}[t]
\vspace{-.15in}
\caption{Ablation studies on:
(a) Video2BEV transformation, the second stage of our method, and synthetic negative samples.
(b) Different training strategies. 
\textbf{Train Together}: we fine-tune the first stage based on the weights pre-trained on ImageNet~\cite{ridnik_imagenet_nips21}, and train the second stage from scratch. 
\textbf{Fine-tune}: we load fine-tuned first-stage weights on UniV, and then train both the first stage and the second stage. 
\textbf{Freeze}: we load fine-tuned first-stage weights on UniV, then fix the first-stage weights and only train the second stage from scratch. 
Notably, the \textbf{Freeze} strategy yields the best results.
(c) Re-ranking different top-k samples in the second stage of our method. Considering the balance between performance and testing time, we choose to re-rank top-32 samples.
D and S denote Drone and Satellite, respectively. 
}
\vspace{-.15in}
\begin{subtable}{.6\linewidth}
    \centering 
    \hfill
    \caption{}
    \resizebox{\textwidth}{!}{\begin{tabular}{c|ccc|cc|cc}
    \shline
    \multirow{2}[2]{*}{\raisebox{0.5ex}{Method}} & Video2BEV   & Second & Synthetic & \multicolumn{2}{c|}{D$\rightarrow$S} & \multicolumn{2}{c}{S$\rightarrow$D} \\
          & transformation &  stage & negatives & R@1 & AP    & R@1 & AP \\
    \hline
    Baseline & \textcolor{red}{\ding{55}}     & \textcolor{red}{\ding{55}}     & \textcolor{red}{\ding{55}}     & 89.87    & 91.28    & 90.01    & 91.36 \\
    BEVs   & \textcolor{nvidiagreen}{\ding{51}}     & \textcolor{red}{\ding{55}}     & \textcolor{red}{\ding{55}}     & 95.01    & 95.64    & 93.44    & 94.44 \\
    Two Stage & \textcolor{nvidiagreen}{\ding{51}}     & \textcolor{nvidiagreen}{\ding{51}}     & \textcolor{red}{\ding{55}}     & 95.86    & 96.48    & 95.01    & 95.78 \\
    Ours  & \textcolor{nvidiagreen}{\ding{51}}     & \textcolor{nvidiagreen}{\ding{51}}     & \textcolor{nvidiagreen}{\ding{51}}     & \textbf{96.29} & \textbf{96.80} & \textbf{96.01} & \textbf{96.57} \\
    \shline
    \end{tabular}
  \label{tab:ablation}}
  \caption{}
  \centering
  \resizebox{\textwidth}{!}{\begin{tabular}{c|ccc|cc|cc}
    \shline
    \multirow{2}[2]{*}{\raisebox{0.5ex}{Strategy}} & Load fine-tuned  & \multirow{2}[2]{*}{\raisebox{0.5ex}{Train first stage}} & \multirow{2}[2]{*}{\raisebox{0.5ex}{Train second stage}} & \multicolumn{2}{c|}{D$\rightarrow$S} & \multicolumn{2}{c}{S$\rightarrow$D} \\
    & first-stage weights &    &   & R@1   & AP    & R@1   & AP \\
    \hline
    Train Together & \textcolor{red}{\ding{55}} & \textcolor{nvidiagreen}{\ding{51}}  & \textcolor{nvidiagreen}{\ding{51}}   & {74.75} & {79.29} & {82.17} & {85.39}\\
    Fine-tune & \textcolor{nvidiagreen}{\ding{51}}  & \textcolor{nvidiagreen}{\ding{51}}     &  \textcolor{nvidiagreen}{\ding{51}} & 96.29  & \textbf{96.83} & 95.29  & 95.99  \\
    Freeze & \textcolor{nvidiagreen}{\ding{51}}  & \textcolor{red}{\ding{55}}     &  \textcolor{nvidiagreen}{\ding{51}} & \textbf{96.29} & 96.80  & \textbf{96.01} & \textbf{96.57} \\
    \shline
    \end{tabular}%
  \label{tab:training}}%
\end{subtable}
\begin{subtable}{.35\linewidth}
\small
    \centering 
    \hfill
    \caption{}
    \resizebox{.88\textwidth}{!}{\begin{tabular}{c|cc|cc}
    \shline
    \multirow{2}[2]{*}{\raisebox{0.5ex}{Top-K}} & \multicolumn{2}{c|}{D$\rightarrow$S} & \multicolumn{2}{c}{S$\rightarrow$D} \\
          & R@1   & AP    & R@1   & AP \\
    \hline
    8     & 96.01  & 96.52  & 95.58  & 96.10  \\
    16    & 96.01  & 96.51  & 95.72  & 96.25  \\
    32    & {96.29} & 96.80  & \pmb{96.01} & 96.57  \\
    64    & {96.29} & {96.81} & \pmb{96.01} & \pmb{96.60} \\
    128    & \pmb{96.43} & {96.98} & \pmb{96.01} & \pmb{96.60} \\
    256    & \pmb{96.43} & \pmb{96.99} & \pmb{96.01} & \pmb{96.60} \\
    512    & \pmb{96.43} & \pmb{96.99} & \pmb{96.01} & \pmb{96.60} \\
    \shline
    \end{tabular}%
  \label{tab:topk}}%
\end{subtable}
\vspace{-.15in}
\end{table*}

\subsection{Comparisons with Competitive Methods}
\begin{figure}[!t]
    \centering
    \includegraphics[width=0.99\linewidth]{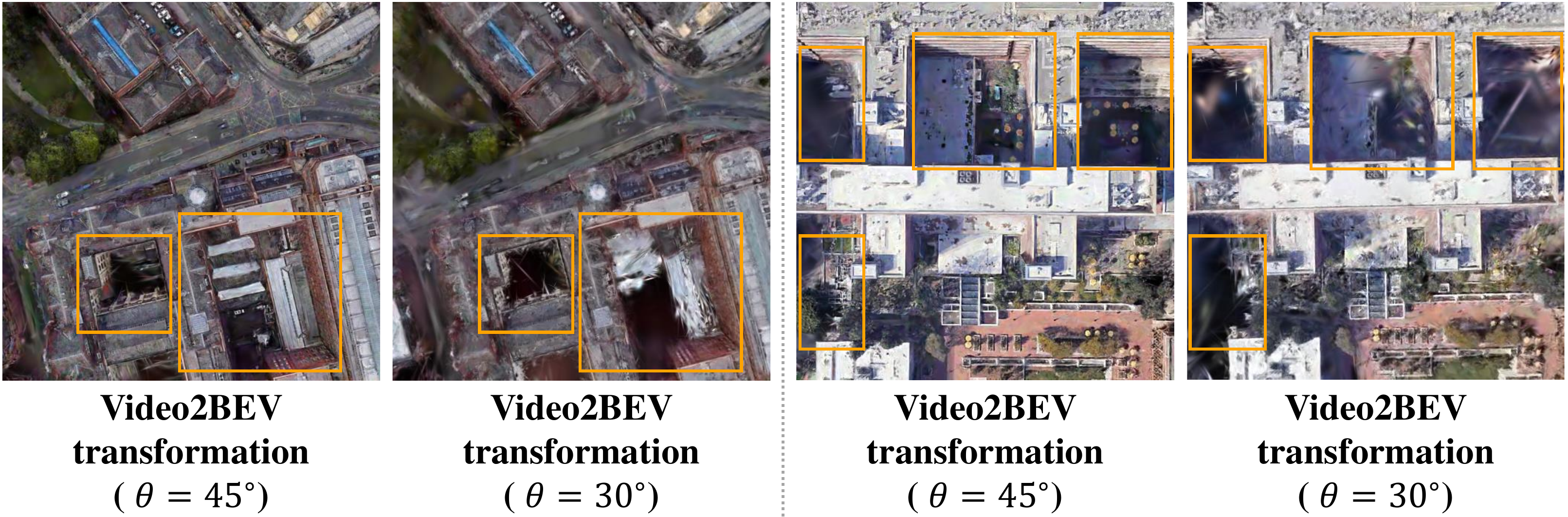}
    \vspace{-0.10in}
    \caption{The transformed BEV comparison of videos with different evaluation $\theta$. We highlight the \textcolor{orangeyellow}{challenging regions}.}
    \label{fig:45_30_bev}
    \vspace{-0.1in}
\end{figure}

{\noindent\textbf{{Quantitative Results}.}}
As shown in \reftab{tab:performance},  we compare the proposed method with other competitive methods on the UniV dataset.
The performance of our method has surpassed that of other competitive methods~\cite{wang_lpn_tcsvt21,dai_fsra_tcsvt21,wang_dwdr_arxiv22,deuser_sample4geo_iccv23}.
On the $45^\circ$ subset, our method achieves gains of 0.30\% Recall and 0.58\% AP for satellite $\rightarrow$ drone compared to the second-best method. 
On the $30^\circ$ subset, all methods experience a performance drop. 
As shown in~\reffig{fig:45_30_bev}, we highlight some imperfect reconstructed regions by the Video2BEV transformation. The lower elevation of the drone flights raises more occlusions (see~\reftab{fig:dataset}), which also compromises our Video2BEV transformation.
Compared to the second best method, our method is still robust, receiving improvements of 3.2\% AP for drone $\rightarrow$ satellite and 5.8\% AP for satellite $\rightarrow$ drone, respectively (see~\reftab{tab:performance}).
All methods are compared in the video setting, which means we temporally average the outputs of frames in a video from the drone view.
For methods with officially released weights (Sample4Geo~\cite{deuser_sample4geo_iccv23}, DWDR~\cite{wang_dwdr_arxiv22}), we test these methods on the $45^\circ$ test set directly and subsequently retrain and evaluate these methods on the $30^\circ$ subsets. 
For methods without official weights (LPN~\cite{wang_lpn_tcsvt21}, FSRA~\cite{dai_fsra_tcsvt21}), we retrain them on both the $45^\circ$ and $30^\circ$ subsets to ensure a fair comparison.

\noindent\textbf{Out-of-Distribution (OOD) Scalability.} We also evaluate the model trained on the UniV dataset ($45^\circ$) on the unseen SUES-200 $45^\circ$~test set in an OOD manner.
SUES-200 dataset contains dense frames collected in real-world environments, including real-world light, shadow transformations, and disturbances.
We observe that our method shows strong OOD potential, surpassing the runner-up method by more than 7\% AP for drone $\rightarrow$ satellite (see~\reftab{tab:realworld}). More results on robustness against weather and other real-world variants can be found in the \textbf{Suppl.}

\begin{figure}[!t]
    \centering
    \includegraphics[width=0.99\linewidth]{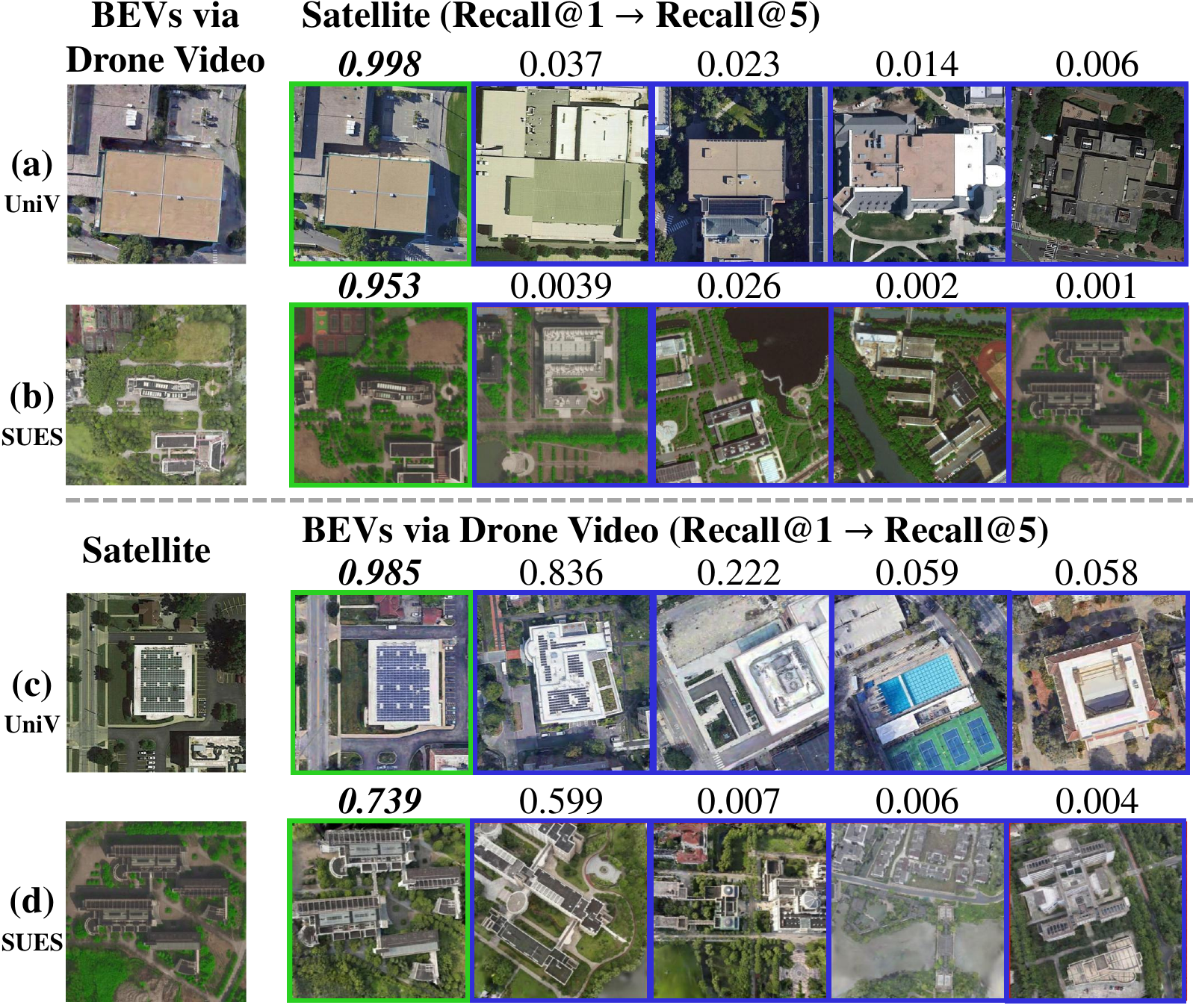}
    \vspace{-0.10in}
    \caption{
    Qualitative results on the UniV (a, c) and SUES-200 (b,d) dataset for Drone $\rightarrow$ Satellite and Satellite $\rightarrow$ Drone. We depict the transformed output (BEVs) as the query or gallery.
    Given queries (left) from different platforms, matched galleries are in \textcolor{nvidiagreen}{green} box, and mismatched galleries are in \textcolor{blue}{blue} box. The scores on the top are similarity scores estimated by the proposed method.
    }
    \label{fig:qualitive_results}
    \label{fig:sues-200}
    \vspace{-.2in}
\end{figure}

\noindent\textbf{Qualitative Results.}
We show qualitative results of the drone geo-localization on the UniV and SUES-200 datasets (see~\reffig{fig:sues-200}).
In our method, drone-view videos are transformed to BEVs by the proposed Video2BEV transformation and we choose the representative sample from the BEV sequence for visualizations.
For drone $\rightarrow$ satellite, we observe that the proposed method effectively retrieves reasonable locations with similar structural features, such as cross-shaped roofs and roofs equipped with solar panels.
For satellite $\rightarrow$ drone, we find a similar result. Our method successfully {retrieves true-matched results at the top of the candidate list} 
among images with similar contents. {We add more visualizations in the supplementary material.}

\subsection{Ablation Study and Further Discussion}
\noindent\textbf{Effect of Primary Components.}
We conduct ablation studies on the UniV dataset ($45^\circ$ subset).
We employ the first stage of our method as the baseline (\textbf{Baseline}), which consists of a shared backbone supervised with the instance loss and contrastive loss. The input data for the baseline are drone-view videos and satellite-view images.
Then, we transform drone-view videos to BEVs via the proposed Video2BEV transformation and adopt BEVs as input for the baseline, denoted as \textbf{BEV}.
Next, we introduce the second stage of our method to the baseline, which is supervised by the matching loss, denoting \textbf{Two Stage}. The negative samples for this architecture are from in-batch samples~\cite{li_albef_nips21}.
Finally, we incorporate the synthetic negative samples in ~\cref{sec:DDGS} to train the second stage of our method and form the final version of our method, referred to as \textbf{Ours}.
As shown in \reftab{tab:ablation}, BEVs receive the largest performance improvement. We attribute this improvement to the reduction of the appearance gap between the drone-view images and the satellite-view images through the proposed Video2BEV transformation.
Additionally, synthetic negative samples contribute to a substantial performance boost due to the enhanced quality of the negative samples for the second stage of Ours.
The two-stage method (Two Stage) also receives improved performance, indicating that many false negative predictions are ranked within the range of the top 32. A fine-grained re-ranking can effectively rectify the matching results from the first stage of our method.

\noindent\textbf{Effect of Training Strategies.}
We explore three different strategies for training.
For the \textbf{Train Together} strategy, we load the matched weights pre-trained on the {ImageNet} dataset~\cite{ridnik_imagenet_nips21}. Then we fine-tune the first stage of the proposed model and train the second stage of the proposed model from scratch.
The \textbf{Fine-tune} strategy entails loading fine-tuned weights of the first stage on the UniV dataset. After this, we fine-tune the first stage with a smaller learning rate while training the second stage from scratch. 
The \textbf{Freeze} strategy consists of loading fine-tuned weights of the first stage on UniV, then fixing all weights of the first stage, while training the second stage from scratch.
The results of three training strategies are in~\reftab{tab:training}.
The \textbf{Train Together} strategy yields the worst results. We attribute this to the difficulty of training both stages simultaneously, as the first stage of the proposed model is designed for coarse-grained retrieval, while the second stage of the proposed model focuses on fine-grained retrieval, relying on the output of the first stage. When both stages are trained together, the first stage fails to retrieve reliable candidates for the second stage, affecting the overall training process.
The \textbf{Fine-tune} strategy achieves a significant performance boost, as the first stage is able to produce reliable embeddings for the second stage.
Finally, we freeze the first stage after loading its corresponding weight. The \textbf{Freeze} strategy yields the best result, and we adopt this strategy.

\noindent\textbf{Effect of Re-ranking Top-K Samples.}
During the test stage, we select top-k samples from the gallery, leveraging the similarity score from the first stage of our method and subsequently re-rank these samples by the second stage. We conduct hyper-parameter experiments with varying values of top-k, and select k $\in$ $\{8, 16, 32, 64, 128, 256, 512\}$ {(see ~\reftab{tab:topk})}. 
Re-ranking the top-512 and top-256 samples yields the best performance and re-ranking the top-256, top-128, top-64, and top-32 samples results in a slight performance drop, respectively. 
Re-ranking the top-16 and top-8 samples leads to a further decline in performance.
Considering the balance between the performance and the testing time, we re-rank the top 32 samples as default.

\section{Conclusion}
In this work, we propose to leverage videos to mitigate the impact of environmental constraints in drone visual geo-localization. 
We propose a new Video2BEV paradigm that transforms drone-view videos into Bird's Eye View (BEV) images by 3D gaussian splatting.
This transformation effectively reduces the \textbf{inter-platform} viewpoint disparity between the drone view and the satellite view. 
Our Video2BEV paradigm also includes a diffusion-based module to generate negative samples, enhancing the  \textbf{intra-platform} discriminative ability of the model.
To support the video setting and validate the proposed framework, we introduce the UniV dataset, a new video-based drone geo-localization dataset.
The dataset includes flight paths of the drone at $30^\circ$ and $45^\circ$ elevation angles and corresponding videos recorded at up to 10 frames per second.
Extensive experiment validates that our Video2BEV paradigm outperforms other competitive approaches in both supervised setting on UniV and OOD testing on unseen SUES-200.
\section{Acknowledgment}
We acknowledge support from 
Guangdong Basic and Applied Basic Research Foundation 2025A1515012281, Nanjing Municipal Science and Technology Bureau 202401035, University of Macau MYRG-GRG2024-00077-FST-UMDF, 
and 
National Key R\&D Program of China (2022ZD0115502), National Natural Science Foundation of China (NO.62461160308, U23B2010), “Pioneer” and “Leading Goose” R\&D Program of Zhejiang (No. 2024C01161).

{
    \small
    \bibliographystyle{ieeenat_fullname}
    \bibliography{main}
}

\input{main-supp}
\end{document}

%% file: main-supp.tex




%
\definecolor{iccvblue}{rgb}{0.21,0.49,0.74}
\def\paperID{2492} 
\def\confName{ICCV}
\def\confYear{2025}



\maketitlesupplementary
\setcounter{table}{3}  
\setcounter{figure}{7}  

\noindent\textbf{Outline.} This supplementary material includes 4 aspects:
\begin{enumerate}
    \item Visualization:
    \begin{itemize}
        \item more visualizations of the Video2BEV transformation:
        \begin{itemize}
            \item comparisons of the Video2BEV transformation at different elevation angles on the UniV;
            \item visualizations of drone-view videos, BEVs, and satellite-view images on the UniV.
        \end{itemize}
        \item more visualizations of the UniV dataset;
        \item more visualizations of synthetic negative samples on the UniV dataset.
    \end{itemize}
    \item Out-of-Distribution scalability test in rainy weather.
    \item Failure case analysis.
    \item Implementation details.
    \item Inference efficiency.
    \item Additional ablation study:
    \begin{itemize}
            \item ablation study for loss weights;
            \item ablation study for FPS;
            \item visualizations of different trajectory lengths.
    \end{itemize}
\end{enumerate}
\section{Visualization}
\subsection{Visualizations of the Video2BEV Transformation}
\begin{figure}[!h]
    \centering
    \includegraphics[width=\linewidth]{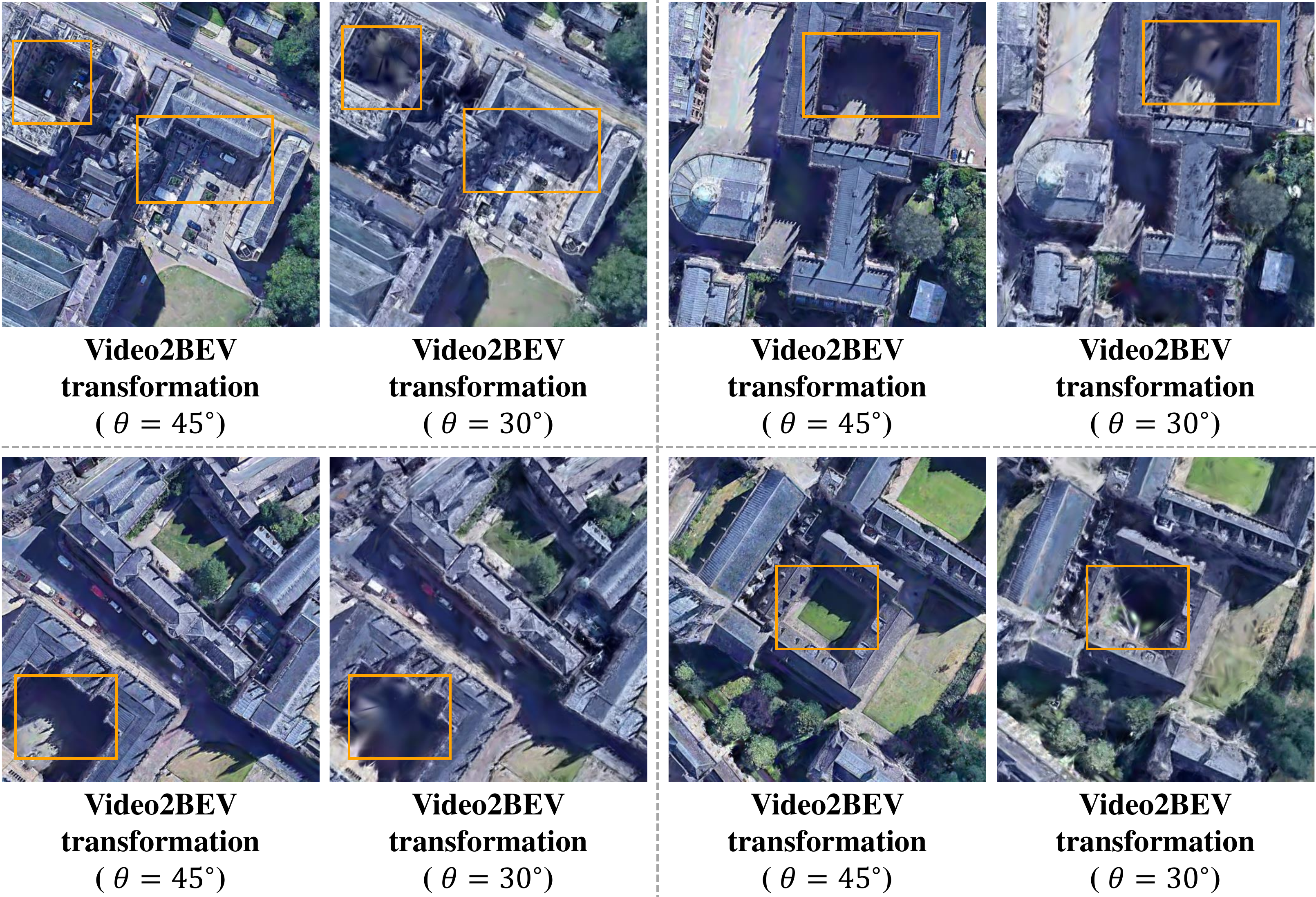}
    \caption{The transformed BEV comparison of videos with different evaluation $\theta$ on the UniV. We highlight the challenging regions.}
    \label{fig:45_30_bev_supp}
\end{figure}
\begin{figure}[!h]
    \centering
    \includegraphics[width=\linewidth]{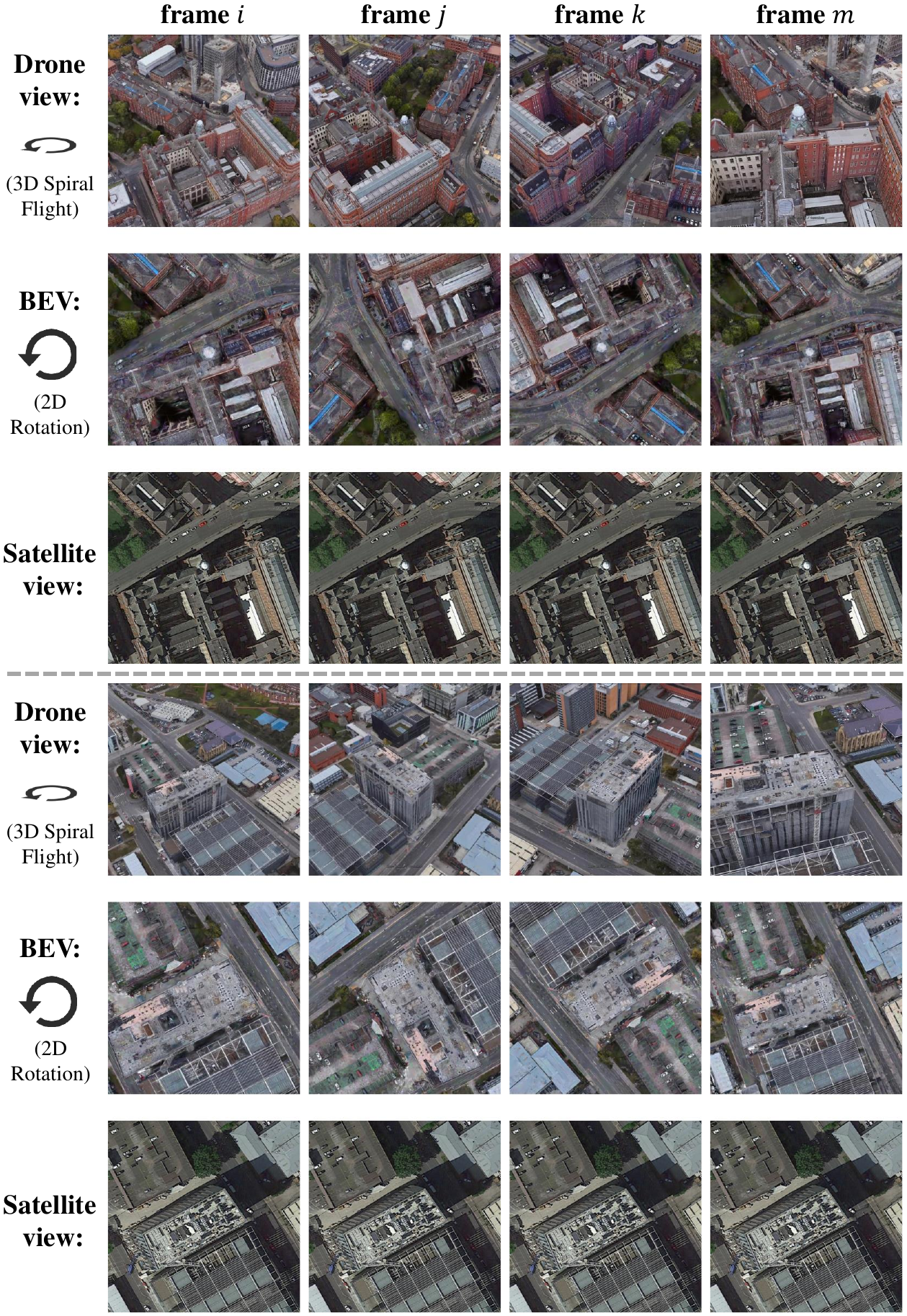}
    \caption{Visualizations of drone-view videos, BEV videos, and satellite-view images on the UniV dataset. $i, j, k, m$ are frame indices, and $i < j < k < m$.}
    \label{fig:drone_bev_satellite}
    \vspace{-0.2in}
\end{figure}
\begin{figure*}[!t]
    \centering
    \includegraphics[width=\linewidth]{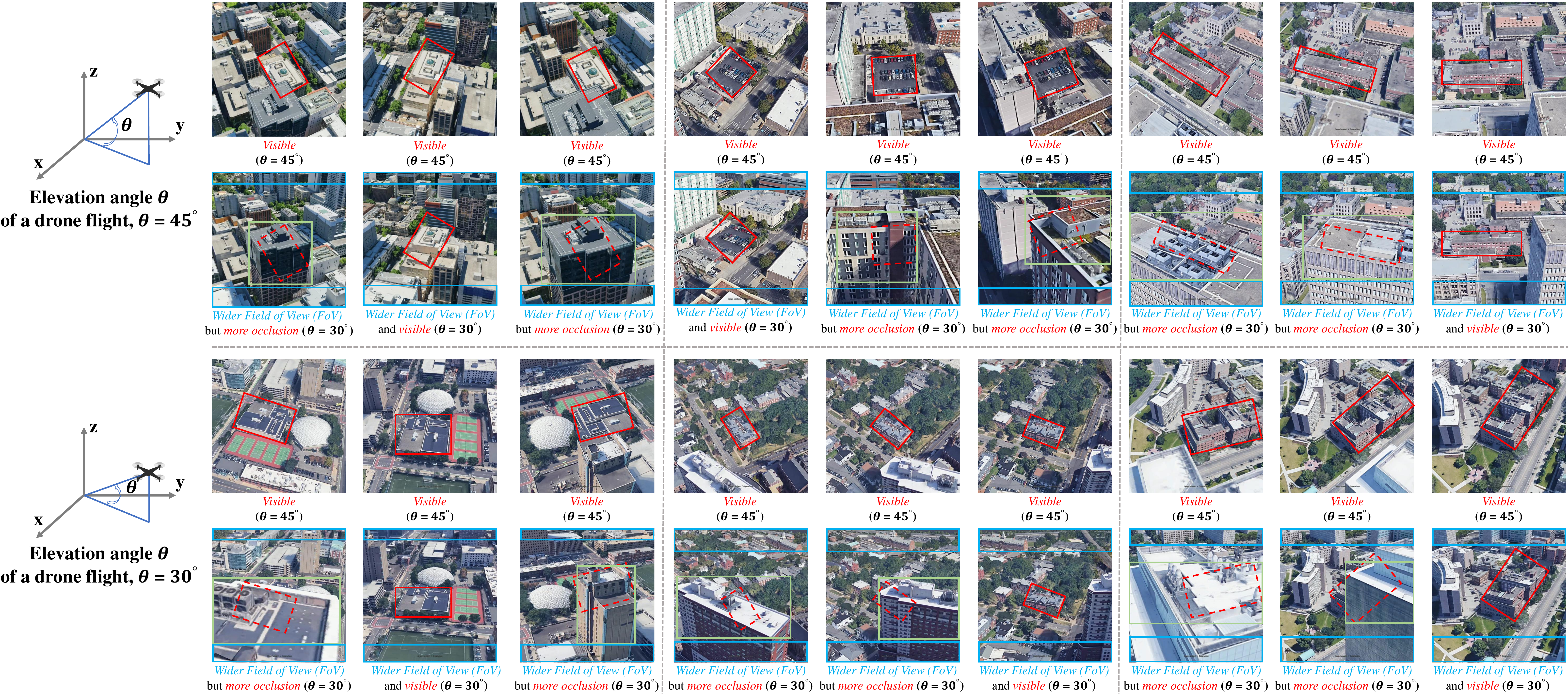}
    \caption{Elevation angles $\theta$ illustration on the UniV dataset. For each case, top row shows $\theta$ = $45^\circ$ and bottom row displays $\theta$ = $30^\circ$. With a lower elevation angle, the new flight captures the target building with \textit{\textcolor[RGB]{0,176,240}{wider Field of View (FoV)}} {but more} \textit{\textcolor{red}{occlusions}}, thereby posing more challenges for drone geo-localization.}
    \label{fig:45_30}
    \vspace{-0.1in}
\end{figure*}
\noindent\textbf{Visualizations of the Video2BEV transformation at different elevation angles.} 
Compared to the $45^\circ$ subset, the $30^\circ$ subset of the UniV dataset presents more occluded cases. We analyze the impact of occlusions and other environmental constraints on the proposed Video2BEV transformation.
As shown in~\reffig{fig:45_30_bev_supp}, the proposed Video2BEV transformation produces satisfactory BEVs at a $45^\circ$ elevation angle, especially in areas between tall buildings.
At the $30^\circ$ elevation angle, some regions reconstructed by the Video2BEV transformation exhibit imperfections. These imperfect regions are primarily located between buildings, where it is challenging for drones to capture clear images at a relatively low elevation angle. 
Despite the imperfectly reconstructed regions, the proposed Video2BEV transformation significantly narrows the disparity between the drone view and the satellite view.

\noindent\textbf{Visualizations of Drone-view Videos, BEVs, and Satellite-view Images.} 
We provide visualizations of images from different platforms. For each building, both drone-view and Bird’s Eye View (BEV) data are in video format and satellite-view data is in image format. 
Specifically, drones follow a spiral path around the target building, completing three circular flights.
For BEVs, in the training set, we incorporate rotation angles and varying heights into BEV camera poses, generating a sequence of rotating and scaled-down BEVs (see~\reffig{fig:drone_bev_satellite}). The 2D rotation is designed for data augmentation.
In the test set, we only increase the height of the BEV camera poses and render a sequence of scaled-down BEV images. 
The satellite view contains one image for each location.
After the proposed Video2BEV transformation, the BEVs align with the same viewing direction as the satellite view and exhibit a similar color pattern to the drone view.

\subsection{Visualizations of the UniV Dataset}
We provide additional visualizations of $45^\circ$ and $30^\circ$ elevation angles in the UniV dataset (see~\reffig{fig:45_30}). Although both videos capture the same building, they differ significantly between the two elevation angles.
Videos captured at a $45^\circ$ elevation angle provide overall views of the core areas of the target building, with these areas visible in most cases. In contrast, at the relatively lower elevation angle of $30^\circ$, drone-view videos offer a wider field of view but also introduce more occlusions. Consequently, core areas of the target building are occluded in some frames while remaining visible in others (see~\reffig{fig:45_30}), effectively simulating outputs from real-world drone flights. More visualizations can be found in the \textbf{\emph{UniV-dataset.mp4}}.
\subsection{Visualizations of Synthetic Negative Samples}
\begin{figure}[!t]
    \centering
    \includegraphics[width=\linewidth]{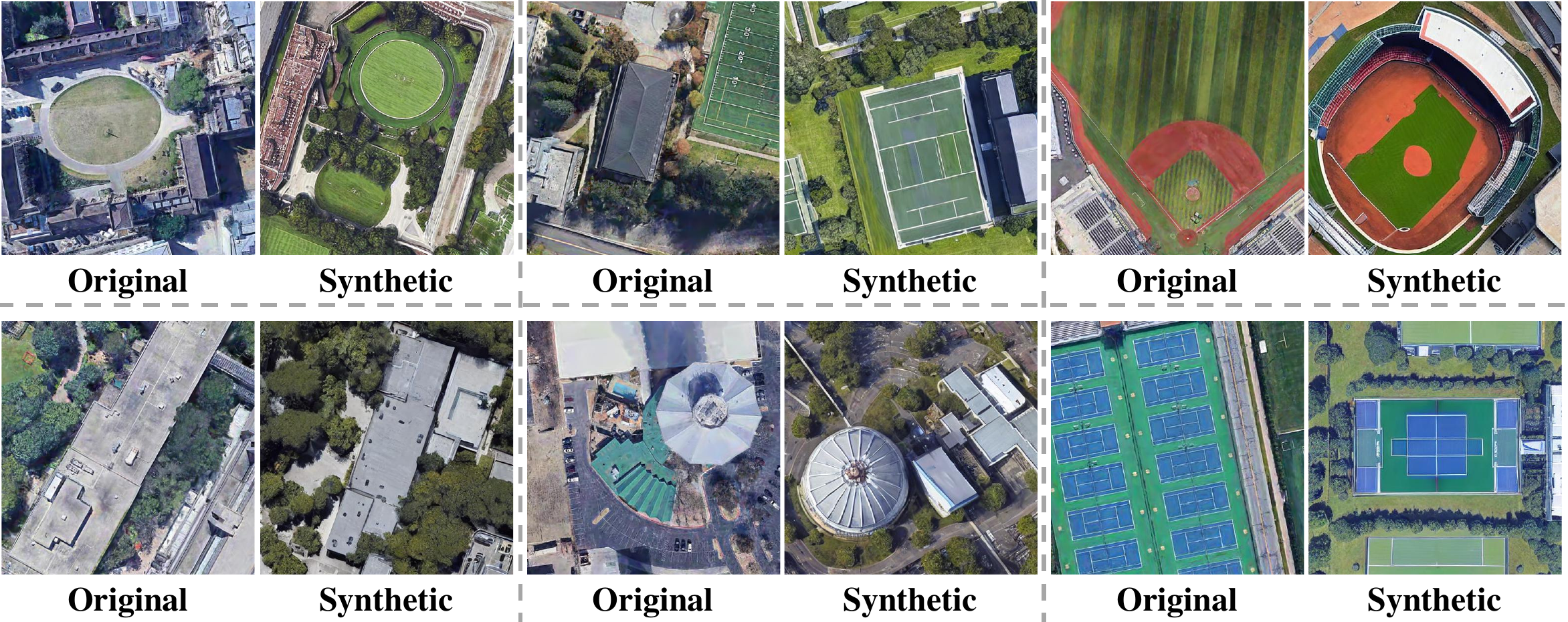}
    \caption{Visualizations of original images and synthetic hard negative samples on the UniV dataset.}
    \label{fig:synthetic_samples}
    \vspace{-0.1in}
\end{figure}
We provide additional visualizations of original and synthetic images (see~\reffig{fig:synthetic_samples}). 
Synthetic negative samples exhibit similarities to the original samples in terms of the architectural features and color patterns of the buildings. 
For cases in the first row, the synthetic samples have architectural features resembling the original images, such as the circular lawn, the green sports field, and the oval stadium.
In the second row, the synthetic samples exhibit similar color patterns to those of the original images.
Despite the similarities, the architectural details differ between the original and synthetic samples, making the synthetic samples suitable for serving as negative samples.
\begin{figure}[!t]
    \centering
    \includegraphics[width=\linewidth]{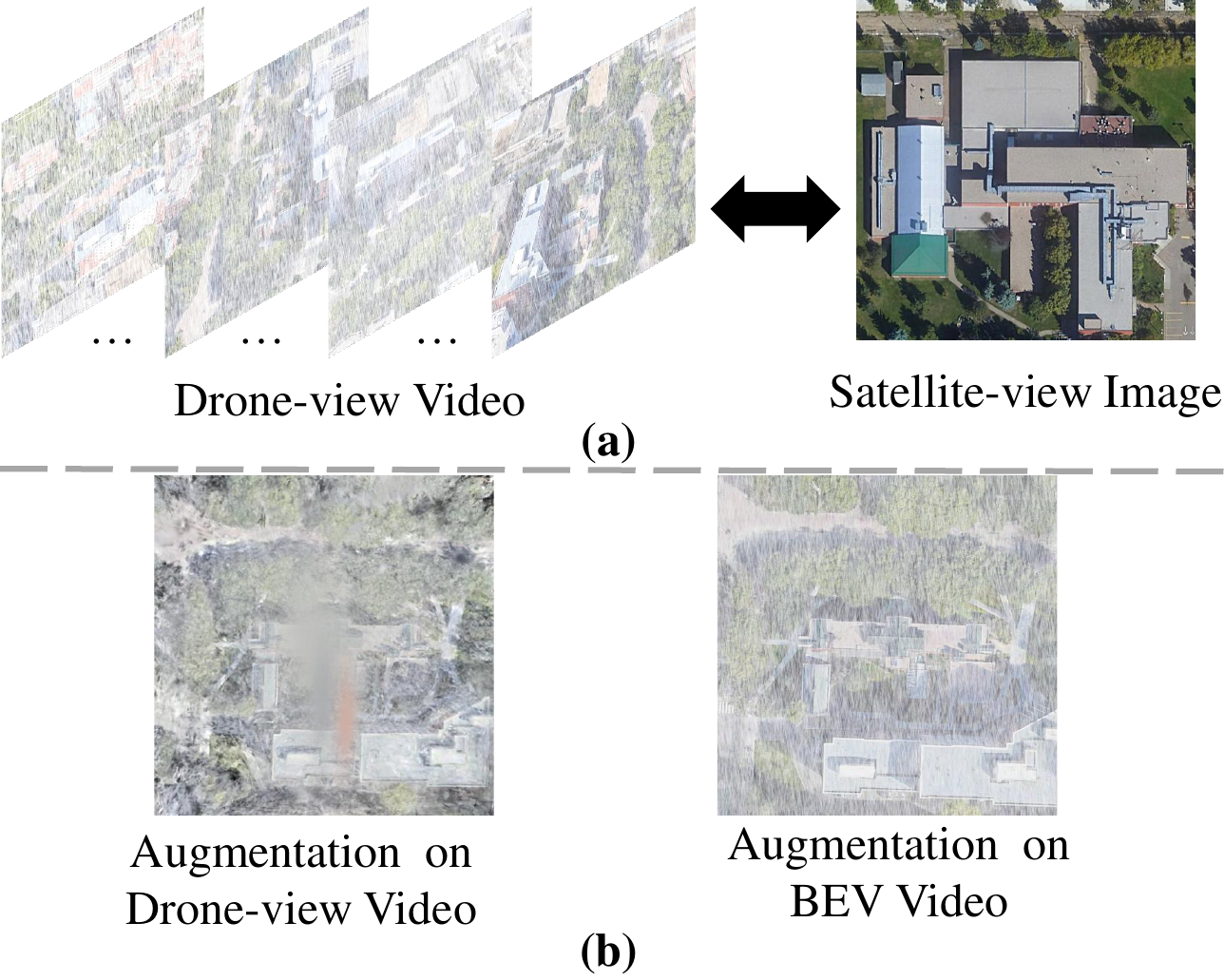}
    \caption{{Visualizations of \textbf{(a) video-based geo-localization under rainy weather setting} and \textbf{(b) comparisons of rainy data augmentations}.}}
    \label{fig:rainy_weather}
    \vspace{-0.1in}
\end{figure}
\section{{Out-of-Distribution (OOD) scalability test in rainy weather}}
Here, we propose a new research direction, namely, video-based geo-localization under rainy weather. Specifically, this setting aims to match drone-view videos (in rainy weather) with geo-tagged satellite images (in clean weather). See \reffig{fig:rainy_weather}\textcolor{cvprblue}{a} for more details.  
We perform rainy data augmentation and out-of-distribution (OOD) testing on the $45^\circ$ \& 2 fps test set (see \reftab{tab:rainy_weather}).  
Compared to Tab.~\textcolor{cvprblue}{2a}, all methods, including ours, show suboptimal performance, indicating that this setting is a challenging one.  
We explore two types of combinations of rainy data augmentation and the Video2BEV transformation in our method.  
The first type applies rainy data augmentation to the clean drone-view videos (Ours$^\dagger$ in \reftab{tab:rainy_weather}), meaning the augmentation is applied \textbf{before} the Video2BEV transformation.  
The second type applies rainy data augmentation to the clean BEV videos (Ours$^*$ in \reftab{tab:rainy_weather}), meaning the augmentation is applied \textbf{after} the Video2BEV transformation.  
The performance of these two combinations varies, which we attribute to the sensitivity of the vanilla version of 3DGS to the input in the Video2BEV transformation. Predicting accurate camera poses and point clouds from rainy videos remains challenging, leading to the imperfect visual quality of BEVs from rainy inputs (see \reffig{fig:rainy_weather}\textcolor{cvprblue}{b}). This finally hinders accurate matching.  
However, we are confident that future advances in efficient 3DGS may help solve this problem. We leave this as a future research direction.

\begin{table}[!t]
  \centering
  \caption{{Comparisons in terms of an out-of-distribution testing on the rainy weather setting ($45^\circ$ test set). $\dagger$ and $*$ denote rainy data augmentation on drone-view videos and BEV-view videos, respectively.}}
    \resizebox{.46\textwidth}{!}{\begin{tabular}{c|cc|cc}
    \shline
    \multirow{2}[4]{*}{\raisebox{1.5ex}{Method}} & \multicolumn{2}{c|}{D$\rightarrow$S} & \multicolumn{2}{c}{S$\rightarrow$D} \\
\cline{2-5}          & R@1    & AP    & R@1    & AP \\
    \hline
    LPN   & 2.28  & 3.58  & 24.82  & 29.64  \\
    FSRA  & 16.83  & 21.18  & 44.08  & 49.42  \\
    DWDR  & 33.67  & 38.91  & 55.63  & 60.85  \\
    Sample4Geo & 60.48  & 64.62  & 72.32  & 75.71  \\
    Ours$^\dagger$ & 30.10  & 33.75  & 36.80  & 40.98  \\
    Ours$^*$ & 65.05  & 67.49  & 83.02  & 84.83  \\
    \shline
    \end{tabular}}
  \label{tab:rainy_weather}%
  \vspace{-0.1in}
\end{table}%
\begin{figure}[!t]
    \centering
    \begin{subfigure}{\linewidth}
        \centering
        \includegraphics[width=\linewidth]{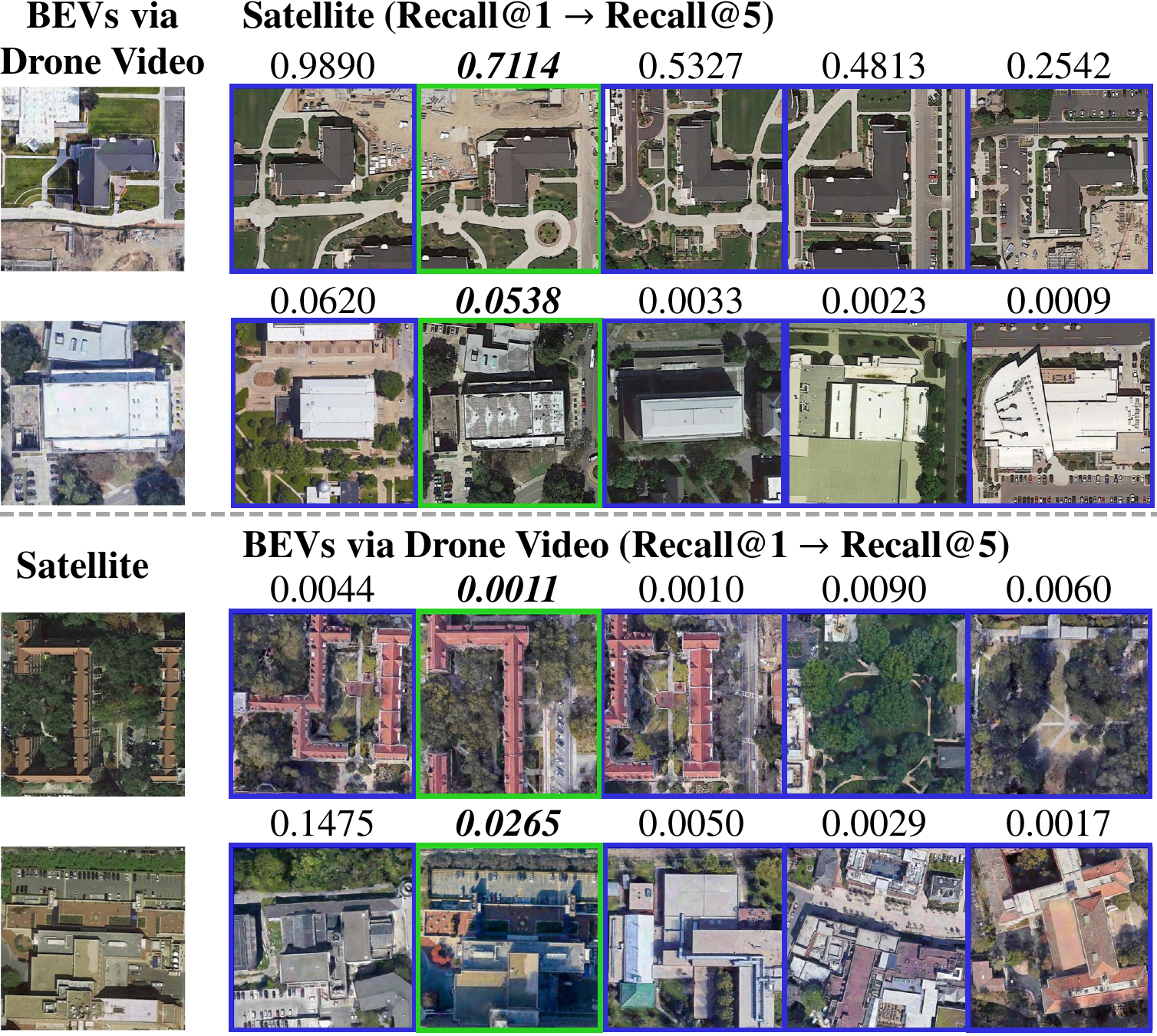}
        \caption{}
        \label{fig:failure_cases}
    \end{subfigure}
    \vspace{0.5em} 
    \begin{subfigure}{\linewidth}
        \centering
        \includegraphics[width=\linewidth]{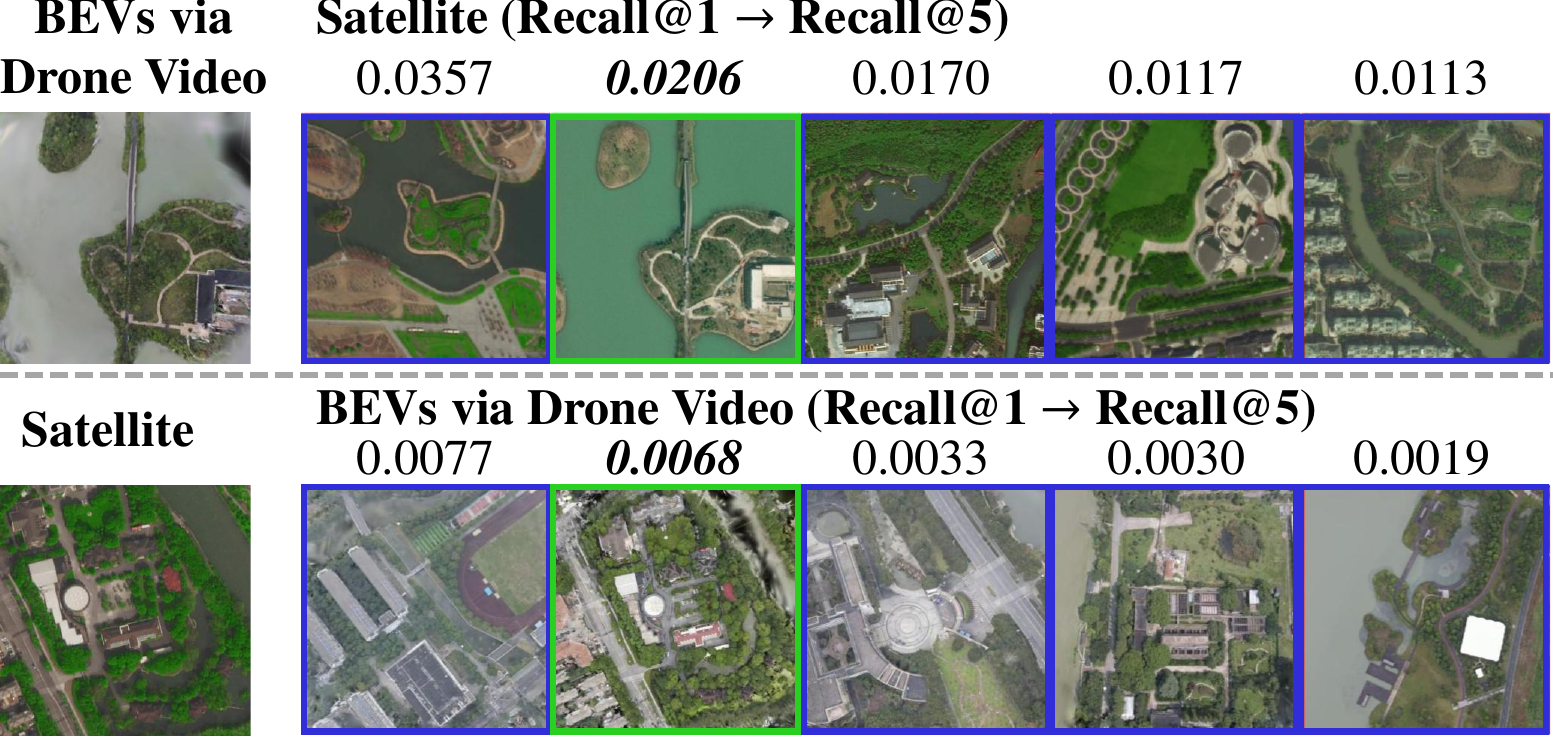}
        \caption{}
        \label{fig:sues200}
    \end{subfigure}
    \vspace{-0.35in}
    \caption{{Typical failure cases} for Drone $\rightarrow$ Satellite and Satellite $\rightarrow$ Drone on the UniV (a) and SUES-200 (b) datasets. 
    {We observe that the failures are mainly due to two factors. 
    First, some buildings were under construction, which is quite different from the current view. 
    Second, some satellite-view photo color is not accurate, and some similar buildings are false-matched. } 
        Given queries (left) from different platforms, matched galleries are in \textcolor{nvidiagreen}{green} box, and mismatched galleries are in \textcolor{blue}{blue} box. Scores on the top are similarity scores estimated from our method.}
    \vspace{-0.2in}
\end{figure}

\section{Failure Case Analysis}
We provide additional qualitative visualizations of retrieval results, with a particular focus on the failure cases on the UniV (\reffig{fig:failure_cases}) and SUES-200 (\reffig{fig:sues200}) datasets.
In these cases, the proposed method fails to recall the matched image in top-1.
We observe that it is challenging because the recalled top-1 image has a very similar pattern to the query image, particularly in terms of the appearance and structure of the geographic target in the two images.
In the first case, all recalled images share a similar structure, and the predicted scores are relatively high.
In the second case, all recalled images have a white rectangular roof. The roof of the ground truth image turns partly gray, which affects the retrieval prediction of our method.
In the third case, recalled top-3 results have similar red roofs, making it challenging for the proposed method to accurately retrieve the ground truth building.
In the fourth case, the recalled top-1 image has a similar architectural style to the query, and the ground truth image is in shadow. Both factors contribute to an inaccurate retrieval result.
In the last two cases, satellite-view photo color is not consistent with that of drone-view, resulting in false positive results.

\section{Implementation details}
\noindent\textbf{Data processing.} 
(\textbf{i}) Video2BEV transformation. We estimate camera poses from drone-view videos using 8 NVIDIA GeForce RTX 4090 GPUs. The subsequent 3DGS training and BEV rendering are also conducted using the same computing resources. It takes less than 1 second to render 50 images with 512 $\times$ 512 with vanilla 3DGS.
(\textbf{ii}) Hard Negative Sample Synthesis. It consists of caption generation, fine-tuning of the Stable Diffusion network, and hard sample synthesis, carried out on a single RTX 4090 GPU. 
All data will be released upon acceptance.

\noindent\textbf{Model training.}
We adopt a two-stage training strategy. 
In the first stage, the encoder is based on the ViT architecture and supervised with instance loss and contrastive loss. In the second stage, the encoder is frozen and MLPs are optimized with the matching loss.
All these experiments are conducted on 1 NVIDIA A800.

\begin{table}[!t]
  \centering
  \caption{{Inference efficiency analysis.}}
    \resizebox{0.48\textwidth}{!}{
    \begin{tabular}{c|ccccccc}
        \shline
        Method        & LPN     & FSRA   & DWDR   & Sample4Geo  & Ours \\
        \hline
        Backbone      & ResNet50 & ViT-S  & Swin-B & ConvNeXt-B & ViT-S \\
        \# Params./M  & 211.00   & 197.66 & 332.90 & 334.03     & 215.26 \\
        GFLOPs        & 16.35    & 24.60  & 30.38  & 90.54      & 28.38 \\
        Time/ms       & 6.76     & 3.77   & 14.23  & 10.88      & 6.26 \\
        \shline
    \end{tabular}}%
  \label{tab:efficency}%
  \vspace{-.1in}
\end{table}%
\section{{Inference efficiency}} 
We provide details of two modules and an overall inference efficiency.
(1) Hard negative sample synthesis is only used during training and does not impact the inference.
(2) Our optimized 3DGS~\cite{hanson2025speedy} renders Video2BEV transformation for 0.15s with 1 Nvidia 4090. 
It can be executed on a remote server in advance, thus not affecting on-device latency.
Future advances in efficient 3DGS may facilitate our work to on-device rendering. The rendering efficiency is out-of-the-scope of our work, and we leave it as the future work.
(3) The overall on-device ranking time is in ~\reftab{tab:efficency}.
Ours takes slightly more time per sample than FSRA, but is more efficient than the rest of the competitive methods.
\begin{table}[!t]
    \centering
    \caption{Ablation study for loss weights $\lambda$.}
    \label{tab:loss_balance}
    \vspace{-0.1in} 
    \resizebox{0.4\textwidth}{!}{\begin{tabular}{c|cc|cc} 
      \shline
      & \multicolumn{2}{c|}{D$\rightarrow$S} & \multicolumn{2}{c}{S$\rightarrow$D} \\
      \cline{2-5} \raisebox{1.5ex}{$\lambda$} & {R@1} & {AP} & {R@1} & {AP} \\
      \hline
      0.10  & 94.58  & 95.25  & 92.87  & 93.81  \\
      0.50  & 95.01  & 95.62  & 93.01  & 94.05  \\
      1.00 (Ours)  & \textbf{95.01}  & \textbf{95.64}  & \textbf{93.44}  & \textbf{94.44}  \\
      \shline
    \end{tabular}}
    \vspace{-0.1in}
\end{table}
\begin{table}[!t]
  \centering
  \caption{{Ablation study on the 45$^\circ$ subset for different FPS, \textbf{best} and \ui{second best}. Overall, 5 and 10 FPS outperform a little bit over 2 FPS.}}
  \resizebox{0.48\textwidth}{!}{
    \begin{tabular}{c|cc|cc|cc||cc|cc|cc}
    \shline
    \multirow{3}[6]{*}{\raisebox{3ex}{Method}} & \multicolumn{6}{c||}{D$\rightarrow$S}                       & \multicolumn{6}{c}{S$\rightarrow$D} \\
\cline{2-13}          & \multicolumn{2}{c|}{FPS = 2} & \multicolumn{2}{c|}{FPS = 5} & \multicolumn{2}{c||}{FPS = 10} & \multicolumn{2}{c|}{FPS = 2} & \multicolumn{2}{c|}{FPS = 5} & \multicolumn{2}{c}{FPS = 10} \\
\cline{2-13}          & R@1    & AP    & R@1    & AP    & R@1    & AP    & R@1    & AP    & R@1    & AP    & R@1    & AP \\
    \hline
    LPN   & 86.31  & 88.34  & \ui{86.45}  & \ui{88.47}  & \textbf{86.59}  & \textbf{88.56}  & 83.31  & 85.60  & \ui{84.17}  & \ui{86.26}  & \textbf{84.20}  & \textbf{86.33}  \\
    FSRA  & 88.59  & 90.25  & \ui{88.73}  & \ui{90.36}  & \textbf{88.87}  & \textbf{90.45}  & \ui{87.30}  & \ui{89.17}  & \ui{87.30}  & 89.11  & \textbf{87.59}  & \textbf{89.33}  \\
    DWDR  & \ui{91.73}  & \ui{92.96}  & \textbf{91.87}  & \textbf{93.06}  & \ui{91.73}  & \ui{92.96}  & 89.87  & 91.45  & \textbf{90.01}  & \textbf{91.57}  & \textbf{90.01}  & \ui{91.56}  \\
    Sample4Geo & \textbf{96.29}  & \textbf{96.75}  & \ui{96.14}  & \ui{96.62}  & \ui{96.14}  & 96.61  & \textbf{95.29}  & {95.99}  & \textbf{95.29}  & \textbf{96.00}  & \textbf{95.29}  & \textbf{96.00}  \\
    Ours & \ui{96.29}  & \ui{96.80}  & \textbf{96.43}  & \textbf{96.92}  & {96.15}  & 96.70  & \textbf{96.01}  & \textbf{96.57}  & \ui{95.58}  & \ui{96.21}  & {94.58}  & {95.42}  \\
    \shline
    \end{tabular}}%
  \label{tab:fps}%
\end{table}%
\begin{figure}[!t]
    \centering
    \includegraphics[width=\linewidth]{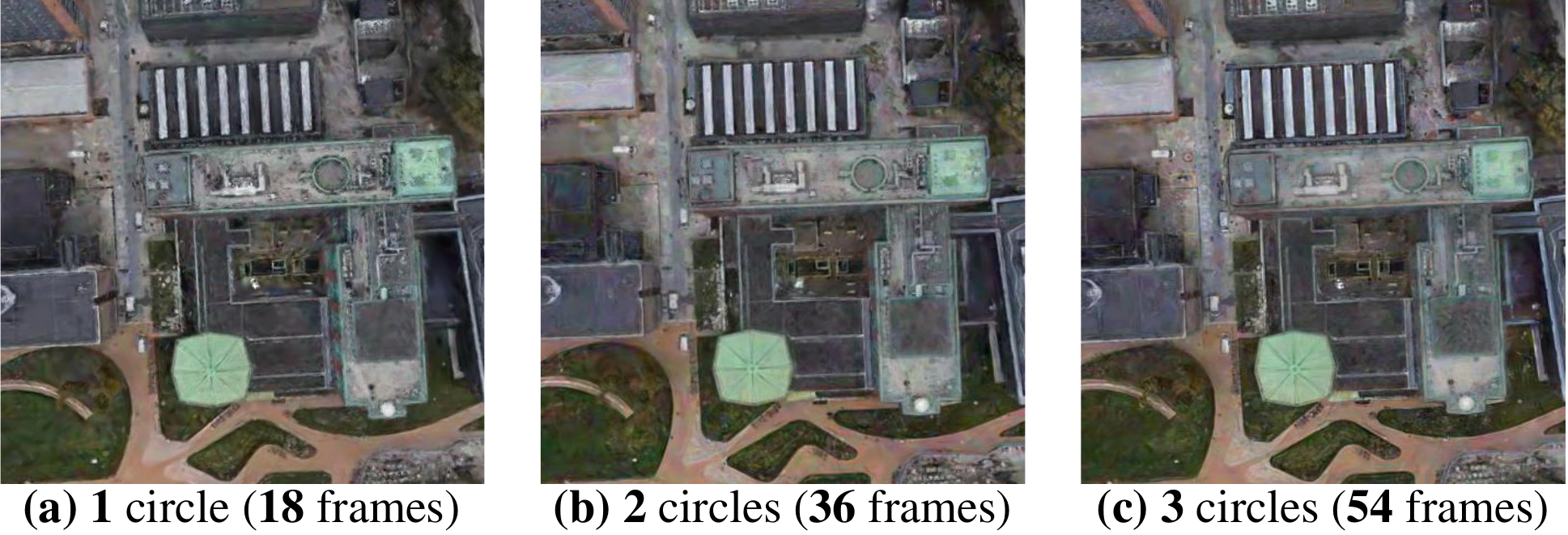}
    \caption{{Visualizations of different trajectory lengths in Video2BEV transformation. We generate BEV by 3DGS via 3 different trajectory lengths. We find 1/3 trajectory (a) providing competitive visual quality.}}
    \label{fig:trajectory_bev}
    \vspace{-0.2in}
\end{figure}
\section{Ablation study}
\subsection{Ablation study for loss weights}
We employ a two-stage training strategy. 
In the first stage, the losses are the instance loss $\mathcal{L}_{I}$ and the contrastive loss $\mathcal{L}_{C}$, resulting in the total loss $\mathcal{L} = \mathcal{L}_{I} + \lambda \mathcal{L}_{C}$. 
The second stage utilizes the matching loss $\mathcal{L}_M$ only. 
By default, we set weights=1 following previous works. 
We add an ablation study on $\lambda$ in the first stage (\reftab{tab:loss_balance}).
We observe that $\lambda=1$ achieves the best performance.
\subsection{{Ablation study for different FPS}}
UniV's necessity lies in \textbf{multi-view video} and \textbf{occlusion-heavy elevation angle}, not specific FPS. 
Previous experiments all focus on the 2-FPS subset in the UniV, and we additionally provide an ablation study for different FPS (see~\reftab{tab:fps}).
Compared to a single image, as long as videos cover more areas, videos offer better robustness to occlusion and viewpoint changes. High FPS is not a key, but it provides more chances to see unoccluded areas. Therefore, 5/10 FPS outperforms a little bit over 2FPS~in~\reftab{tab:fps}. How to effectively and efficiently process high-fps videos is also a future research direction.

\subsection{{Visualizations of different trajectory lengths in Video2BEV transformation}}
We adopt all frames, \ie, 54 frames, in the 2-fps video by default. Here we provide visualizations of different trajectory lengths (see~\reffig{fig:trajectory_bev}). We find 1/3 trajectory, \ie, 18 frames,  is enough to train 3DGS, achieving competitive visual quality.

